\documentclass{article} % For LaTeX2e
\usepackage{iclr2026_conference,times}
\iclrfinalcopy
\usepackage{amsmath,amsfonts,bm}

\def\eqref#1{equation~\ref{#1}}
\def\1{\bm{1}}

\DeclareMathAlphabet{\mathsfit}{\encodingdefault}{\sfdefault}{m}{sl}
\SetMathAlphabet{\mathsfit}{bold}{\encodingdefault}{\sfdefault}{bx}{n}

\usepackage{hyperref}
\usepackage{url}

\usepackage{booktabs}
\usepackage{multirow}
\usepackage{booktabs} 
\usepackage{xcolor}   
\usepackage{array}
\usepackage{makecell}
\usepackage{enumitem}
\usepackage{arydshln}
\usepackage{pifont}
\usepackage{wrapfig}
\usepackage{algorithm}
\usepackage{amsmath}
\usepackage{algpseudocode}
\usepackage[utf8]{inputenc}
\usepackage{amssymb}
\usepackage{tcolorbox}

\usepackage{threeparttable}
\usepackage{bbding}

\newcommand{\no}{\XSolidBrush}
\newcommand{\midmark}{$\bigcirc$} 
\definecolor{ao(english)}{rgb}{0.0, 0.5, 0.0}

\newcommand{\yes}{\CheckmarkBold}

\usepackage{devanagari}

\title{Neuron-Aware Data Selection in Instruction Tuning for Large Language Models}

\author{Xin Chen\textsuperscript{\rm 1,2,5,}\thanks{Equal contribution.} ,
    Junchao Wu\textsuperscript{\rm 1,$\ast$},
    Shu Yang\textsuperscript{\rm 4},
    Runzhe Zhan\textsuperscript{\rm 1},
    Zeyu Wu\textsuperscript{\rm 1},
    Min Yang\textsuperscript{\rm 2,3,$\dagger$},\\
    \textbf{Shujian Huang\textsuperscript{\rm 5},
    Lidia S. Chao\textsuperscript{\rm 1}, 
    Derek F. Wong\textsuperscript{\rm 1,}\thanks{Corresponding author.}} \\
    \textsuperscript{\rm 1}\textnormal{NLP\textsuperscript{2}CT} Lab, Department of Computer and Information Science, University of Macau, \\
    \textsuperscript{\rm 2}Artificial Intelligence Research Institute, Shenzhen University of Advanced Technology, \\
    \textsuperscript{\rm 3}Shenzhen Institutes of Advanced Technology, Chinese Academy of Sciences, \\
    \textsuperscript{\rm 4}Provable Responsible AI and Data Analytics Lab, KAUST, \\
     \textsuperscript{\rm 5}National Key Laboratory for Novel Software Technology, Nanjing University \\
    \texttt{\small nlp2ct.\{xinchen,junchao,runzhe,zeyu\}@gmail.com, min.yang@siat.ac.cn} \\
    \texttt{\small shu.yang@kaust.edu.sa,\{derekfw,lidiasc\}@um.edu.mo} \\
}

\begin{document}

\maketitle

\begin{abstract}
Instruction Tuning (IT) has been proven to be an effective approach to unlock the powerful capabilities of large language models (LLMs). 
Recent studies indicate that excessive IT data can degrade LLMs performance, while carefully selecting a small subset of high-quality IT data can significantly enhance their capabilities. Therefore, identifying the most efficient subset data from the IT dataset to effectively develop either specific or general abilities in LLMs has become a critical challenge.
To address this, we propose a novel and efficient framework called \textsc{Nait}. \textsc{Nait} evaluates the impact of IT data on LLMs performance by analyzing the similarity of neuron activation patterns between the IT dataset and the target domain capability. Specifically, \textsc{Nait} captures neuron activation patterns from in-domain datasets of target domain capabilities to construct reusable and transferable neuron activation features. It then evaluates and selects optimal samples based on the similarity between candidate samples and the expected activation features of the target capabilities.
Experimental results show that training on the 10\% Alpaca-GPT4 IT data subset selected by \textsc{Nait} consistently outperforms methods that rely on external advanced models or uncertainty-based features across various tasks. Our findings also reveal the transferability of neuron activation features across different capabilities of LLMs. In particular, IT data with more logical reasoning and programmatic features possesses strong general transferability, enabling models to develop stronger capabilities across multiple tasks, while a stable core subset of data is sufficient to consistently activate fundamental model capabilities and universally improve performance across diverse tasks. 
\end{abstract}

\section{Introduction}
\label{sec:introduction}

% 大型语言模型（LLMs）的指令调优（Instruction Tuning）已成为提升模型与人类意图对齐的核心技术[1,2]。尽管现有研究表明，精选少量高质量指令数据即可显著提升模型性能（如LIMA[3]仅使用1K数据），但如何在不依赖外部资源的前提下，系统化筛选出能够增强目标能力且保留基础语言知识的数据，仍是一个关键挑战。现有方法多依赖于外部模型（如ChatGPT）或附加数据集进行数据评分[4,5]，这不仅增加计算成本，还因闭源模型的可控性差而限制其广泛应用。

% 当前方法的主要局限性在于：其一，依赖表面特征（如语义相似性）或不确定度指标，难以捕捉训练样本对模型计算机制的实际影响[6]；其二，重复计算大规模数据的特征导致效率低下。针对这些问题，本文提出\textsc{Nait}（基于神经元激活的高效数据选择框架），其核心思想在于利用预训练模型自身的神经元激活模式，构建可迁移的特征库，实现数据影响力度量与高效筛选。具体来说，我们首先进行神经元激活特征提取，通过量化指令数据对模型神经元的激活程度，构建可复用的特征表征库；随后，进行能力驱动的动态选择，基于少量目标任务的示例，通过神经元激活特征相似性匹配筛选可以有效激活和增强模型相关能力的数据。

% 实验表明，在仅使用\textsc{Nait}筛选的10%数据训练后，LLaMA-2等模型在MMLU（常识推理）、GSM8K（数学问题）等12项任务中平均性能提升2.13。基于神经元激活特征的方法还表现出了强大的可解释性，进一步分析发现，\textsc{Nait}倾向于选择神经元激活反应特征越多样和复杂的指令数据，这类数据能有效增强模型的推理能力和在不同下游任务的综合能力。本文的贡献包括：

% 提出首个基于神经元激活模式的指令数据选择框架\textsc{Nait}，为模型能力定向开发提供新范式；

% 揭示神经元激活特征与模型基础能力保留之间的关联机制，并探索了不同下游任务间的迁移能力；

% 开源跨任务神经元特征库及\textsc{Nait}-Alpaca数据集，一个从 Alpaca-GPT4 数据集中选择高质量的 IT 数据集。

% 总的来说，本文工作将模型神经元机理可解释性与高效调优相结合，为模型能力的高效定向开发和增强提供了理论依据与实践工具。

\begin{figure*}[!ht]
    % \vspace{-1em}
    \centering    
    \includegraphics[width=0.92\textwidth, trim=0 0 0 0]
    {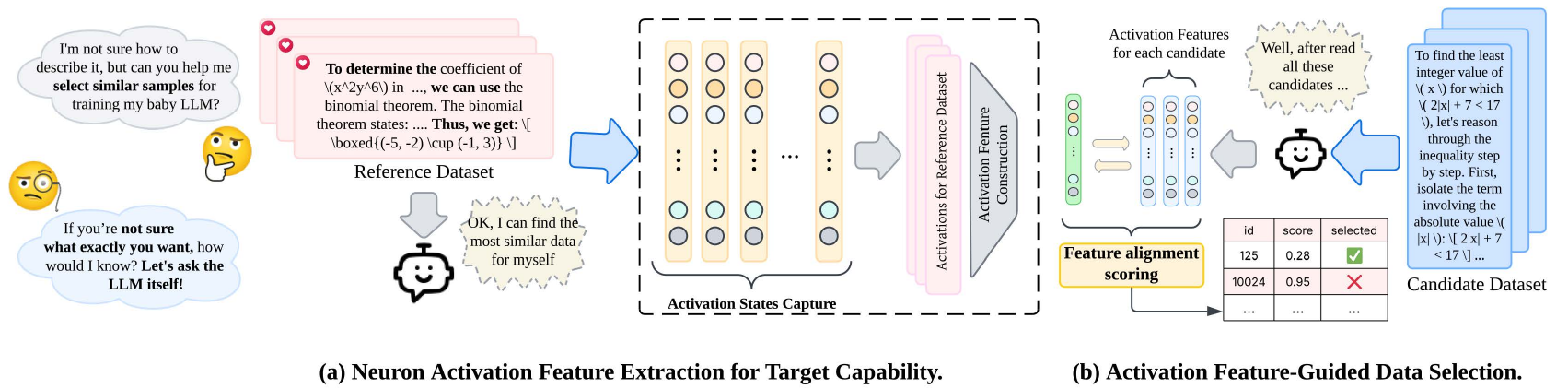}
    \vspace{-1em}
    \caption{\textbf{Overall framework of \textsc{Nait}.} First, we capture the neuron activation of the LLM using in-domain data that we want the model to learn. Next, we construct an activation feature through a dimensionality reduction method. Finally, we evaluate the feature alignment score between this activation feature and the model's activation on each candidate dataset to guide data selection.}
    \label{fig:overall_framework}
    \vspace{-2em}
\end{figure*}
IT of LLMs has become a foundational technique for activating LLMs instruction and knowledge capabilities \citep{DBLP:conf/nips/Ouyang0JAWMZASR22}. Previous studies have shown that excessive
IT data can degrade LLMs performance, while selecting a small amount of high-quality IT data can significantly improve model performance. For instance, LIMA \citep{DBLP:conf/nips/ZhouLX0SMMEYYZG23} achieved impressive results using only 1k IT data. However, a major challenge remains as current approaches lack interpretability in identifying ``high-quality'' data and fail to enhance the specific (one or more) target domain capabilities of LLMs in an open dataset \citep{DBLP:conf/nips/WuGIPG23,DBLP:journals/corr/abs-2408-02085,DBLP:journals/corr/abs-2402-05123}. Moreover, existing SOTA IT data selection methods like Instruction Mining \citep{DBLP:journals/corr/abs-2307-06290}, AlpaGasus \citep{DBLP:conf/iclr/ChenLYWGYTS0HJ24} and SelectIT \citep{DBLP:conf/nips/Liu0W0WH024} often rely on surface-level features, external models and data for scoring, or the model's uncertainty, which are computationally expensive. This limits the scalability of these methods and their broader application to large-scale data.

Inspired by previous works on neuron activations and capabilities of LLMs \citep{DBLP:conf/acl/VoitaFN24,DBLP:journals/corr/abs-2403-11621}, we address these concerns in \textsc{Nait} (\textbf{N}euronal \textbf{A}ctivation-based efficient \textbf{IT} data selection framework), a novel approach designed to evaluate data quality and enhance specific domain capabilities of LLMs by analyzing neuronal activation patterns related to the target domain capabilities of the LLM. The \textsc{Nait} framework is built on the core hypothesis that when an LLM processes a given sample, the closer the model's resulting neuronal activation pattern aligns with the activation pattern of the target capability, the more effectively that sample enhances the LLM's performance in that specific domain.

\textsc{Nait} operates in two main stages. First, we extract neuron activation features corresponding to the LLM's target capability. This process begins by providing a small set of in-domain examples that exemplify the specific capability of interest. As the LLM processes these examples, its neuron activation states are recorded to identify reusable activation features linked to the desired capability. These features serve as a representation of the LLM’s response and neural activation preferences for the target tasks, forming the foundation for the subsequent data selection phase. Next, we perform capability-driven data selection by identifying and selecting IT data that can further activate and enhance the model's relevant capabilities. This is achieved by comparing the neuron activation features of candidate data within an open and diverse dataset (e.g., Alpaca-GPT4) against the in-domain activation features. By prioritizing data that closely matches the neuron activation patterns of the target capability, \textsc{Nait} ensures that the selected data is both relevant and effective for improving the model's specialized abilities. In general, \textsc{Nait} exhibits the following properties: (1) \textbf{Efficiency (\S \ref{sec:main_result} \& \S\ref{sec:ablation} \& \S\ref{sec:cost_efficiency} \& Appendix \ref{sec:IT_for_targeted_domain})}: \textsc{Nait} achieves strong performance with only a small number of in-domain examples and IT data, eliminating the need for external models. It also boasts low computational cost and minimal time requirements. (2) \textbf{Robustness (\S\ref{sec:ablation})}: The framework demonstrates excellent adaptability across different base models, IT datasets, and domain-specific datasets. (3) \textbf{Interpretability (\S \ref{sec:interpretability})}: Qualitative analysis shows that \textsc{Nait} unveils the logical reasoning and programmatic features that possess strong general transferability. Additionally, NAIT identifies a subset of core IT data that remains stable across different task-specific activation features. 
% NAIT identifies an overlapping subset of core IT data that remains stable across different task-specific activation features.

Experimental results demonstrate that training LLaMA-2-7b on only 10\% of the Alpaca-GPT4 IT dataset selected using \textsc{Nait} yields an average performance improvement of 3.24\% across five tasks compared to training on the entire dataset using IT. Further experiments show that NAIT consistently improves performance across different numbers of in-domain data, base models, diverse IT datasets and selection strategies, highlighting NAIT’s strong robustness. The method based on neuron activation features also shows strong interpretability. Further analysis reveals that \textsc{Nait} tends to select IT data with more logical reasoning and programmatic neuron activation features. Such data can effectively enhance the model's reasoning ability and general ability across various downstream tasks. The key contributions of this paper are as follows:

\begin{itemize}[label=\textbullet, topsep=1pt, itemsep=1pt, left=1pt]
    \item Proposing \textsc{Nait}, the first instruction data selection framework based on neuron activation patterns, introducing a new paradigm for the targeted development of model capabilities.
    \item Revealing the correlation between neuron activation features and the retention of fundamental model capabilities, while exploring the transferability of these features across different downstream tasks.
    \item Open-sourcing a cross-task neuron feature library and the Alpaca-\textsc{Nait} dataset, a high-quality IT dataset curated from the Alpaca-GPT4 dataset using our proposed \textsc{Nait}.
\end{itemize}

\section{Related Work}
\label{sec:related_work}
IT plays a crucial role in bridging general pretraining and task-specific alignment, as summarized in Appendix~\ref{sec:it_function}. It enables LLMs to follow instructions, activate latent abilities, and significantly enhance downstream task performance.
\subsection{High-Quality IT Data}
% 高质量指令数据
% 最近的研究已经证实，高质量IT 数据在提升 LLM 偏好对齐和对问题提供准确、相关且无害的响应方面发挥着关键作用~\citep{cite{DBLP:journals/corr/abs-2402-05123,DBLP:conf/naacl/LiuWYCSCYY24}。这些IT数据一般由指令和回答对组成，通过将传统的判别性 NLP 任务数据转化为IT数据的格式（例如 FLAN~\citep{DBLP:conf/iclr/WeiBZGYLDDL22}）或利用手工制作的种子 IT 数据自我指导来进一步合成更多的综合数据（例如 Alpaca~\citep{taori2023alpaca}）。一系列后续研究进一步探索了数据高质量合成机制。例如，Vicuna~\citep{chiang2023vicuna} 通过大规模对话数据增强了 IT 数据的多样性，而 WizardLM~\citep{DBLP:conf/iclr/XuSZG0FTLJ24} 使用 LLM 自动生成具有可控复杂性的 IT 数据集。值得注意的是，LIMA~\citep{DBLP:conf/nips/ZhouLX0SMMEYYZG23} 证明，仅使用 1,000 个精心策划的 IT 数据的 LLM 即可实现与大规模微调相当的性能，从而引入了一种强调数据质量而非数量的新研究范式。这一发现表明 LLM 在预训练阶段已经获得了世界知识，指令调整阶段只需要少量高质量指令数据即可生成高质量响应。然而，涉及IT 数据“高质量”的因素复杂且考虑多维，如何定义“高质量”IT 数据仍然模糊不清，缺乏更具可解释性的选择机制和明确的标准化标准 （\citep{DBLP:journals/corr/abs-2402-05123,DBLP:conf/iclr/0131Z00H24}）。

Recent studies have highlighted the key role of high-quality IT data in enhancing the alignment of LLMs with human preferences as well as generating accurate, relevant, and safe responses~\citep{DBLP:journals/corr/abs-2402-05123,DBLP:conf/naacl/LiuWYCSCYY24}. Typically, IT data consists of instruction-response pairs, which can be created in two main ways: (1) by reformulating traditional NLP task data into the IT format (e.g., FLAN~\citep{DBLP:conf/iclr/WeiBZGYLDDL22} and P3 \citep{DBLP:conf/iclr/SanhWRBSACSRDBX22}) or (2) by synthesizing new data using self-instruction based on a small set of manually crafted seed IT data (e.g., Alpaca~\citep{taori2023alpaca}). Subsequent work has further explored methods for synthesizing high-quality IT data. For example, Vicuna~\citep{chiang2023vicuna} enhanced data diversity with large-scale dialogue data, while WizardLM~\citep{DBLP:conf/iclr/XuSZG0FTLJ24} used LLMs to automatically generate IT datasets with controllable complexity. Notably, LIMA~\citep{DBLP:conf/nips/ZhouLX0SMMEYYZG23} demonstrated that LLMs fine-tuned on only 1k carefully curated IT data can achieve comparable performance of models trained with large-scale dataset, highlighting the importance of data quality over quantity. These findings suggest that pretrained LLMs already contain extensive world knowledge and require only a small amount of high-quality instruction data to yield strong performance in the IT phase. 

However, defining what constitutes ``high-quality'' IT data remains challenging. Data quality is influenced by complex, multidimensional factors, and there is still a lack of interpretable selection mechanisms and standardized criteria~\citep{DBLP:journals/corr/abs-2402-05123}.

% Addressing these gaps is essential for advancing the development of aligned and efficient LLMs.
% \newcommand{\midmark}{\textcolor{blue}{\LARGE$\bullet$}}
\begin{table*}[t]
\centering
\renewcommand{\arraystretch}{1.3}
% \vspace{1em}
\caption{\textbf{Comparison of our method and existing methods.} \yes\ indicates the presence of a feature, \no\ indicates its absence, and \midmark\ denotes partial support.}
\vspace{1em}
\scalebox{0.55}{ % 调整缩放比例
\begin{tabular}{@{}llccccc}
    \toprule
    \textbf{Method}       & \textbf{Feature Source} & \textbf{Externally-Independent} & \textbf{Gradient-Free} & \textbf{Targeted Ability} & \textbf{Cost-Effective} & \textbf{Interpretability} \\ 
    \midrule
    LIMA \citep{DBLP:conf/nips/ZhouLX0SMMEYYZG23} & Human & \yes & \yes & \no & \no & \no \\ 
    Instruction Mining \citep{DBLP:journals/corr/abs-2307-06290} & PPL \& Reward scores \& ... & \yes & \yes & \no & \no & \midmark\\ 
    AlpaGasus \citep{DBLP:conf/iclr/ChenLYWGYTS0HJ24} & ChatGPT Score & \no & \yes & \no & \no & \no \\ 
    SelectIT \citep{DBLP:conf/nips/Liu0W0WH024}  & Multi-granularity Uncertainty  & \yes & \yes & \no & \no & \midmark \\ 
    LESS \citep{DBLP:conf/icml/XiaMGA024}  & Gradient-based  & \yes & \no & \yes & \no & \midmark \\ 
    \textsc{Nait} (Ours)           & Neuron Activation & \yes & \yes & \yes & \yes & \yes  \\ 
    \bottomrule
\end{tabular}
}
\label{tab:methods-comparison}
% \vspace{1em}
\end{table*}

\subsection{Efficient IT Data Selection}

Existing IT data selection strategies typically filter subsets based on metrics such as quality, diversity, or importance~\citep{DBLP:journals/corr/abs-2408-02085}. These strategies fall into four main categories based on how these metrics are extracted: (1) \textit{Handcrafted Feature-Based Methods} rely on manually designed features. For example, DQI~\citep{DBLP:journals/corr/abs-2005-00816} selects high-quality data by leveraging lexical features, \textit{n}-gram frequency, and relational features, while \citet{DBLP:conf/nips/XieS0L23} optimize data distribution using \textit{n}-gram features. Although these methods more interpretable, they are limited to surface-level characteristics and cannot capture deeper data properties; (2) \textit{Model Feature-Based Methods} extract features such as uncertainty or perplexity from model outputs. For instance, Instruction Mining~\citep{DBLP:journals/corr/abs-2307-06290} leverages perplexity and reward scores, and SelectIT~\citep{DBLP:conf/nips/Liu0W0WH024} conducts multi-granularity uncertainty analysis for data selection. However, these methods depend heavily on model outputs, which may introduce bias, and often exhibit ``black-box'' characteristics that limit interpretability. (3) \textit{LLM-as-Scorer Methods} utilize advanced LLMs, such as ChatGPT, to score data based on complex criteria. For instance, InsTag~\citep{DBLP:conf/iclr/LuY0LLTZZ24} evaluates diversity and complexity in IT data, while AlpaGasus~\citep{DBLP:conf/iclr/ChenLYWGYTS0HJ24} scores data quality similarly. These approaches, though effective, are computationally expensive, lack transparency, and are difficult to control or scale due to reliance on closed-source APIs. (4) \textit{Loss and Gradient-based Coreset Sampling Methods}select data based on loss values or gradient information. For example, \citet{DBLP:conf/nips/ChenRBWZSR23} select samples to minimize evaluation loss, and \citet{DBLP:conf/icml/XiaMGA024} use gradient signals to identify critical data points. These methods are computationally intensive, especially for large-scale models, and may suffer from approximation errors.

In summary, existing data selection methods often require substantial computational resources, additional model training, or expensive queries, which limit their scalability to large datasets. Furthermore, they may struggle to enhance specific target domain capabilities of LLMs when applied to open datasets.

\section{Our \textsc{\textsc{Nait}} Method}
\label{sec:our_method}
\subsection{Preliminary}
% Neuron activation analysis offers a novel perspective for understanding the knowledge storage and operational mechanisms of LLMs. Recent studies \citep{DBLP:conf/acl/VoitaFN24,DBLP:conf/emnlp/DurraniSDB20,DBLP:journals/corr/abs-2403-11621} have highlighted the role of neurons in processing knowledge and skills. These studies reveal that the subset of neurons in LLMs is activated by specific knowledge tasks. These task-related neurons are closely associated with the internal knowledge and tasks within LLMs, playing a critical role in the model's ability to learn and solve various knowledge tasks, and even enabling the handling of different linguistic skills \citep{DBLP:conf/emnlp/YuA24,DBLP:conf/emnlp/WangCWSL024,DBLP:conf/acl/TangLH0WZWW24}. 
Neuron activation analysis provides insights into the knowledge storage and operational mechanisms of LLMs \citep{DBLP:conf/acl/VoitaFN24,DBLP:conf/emnlp/DurraniSDB20,DBLP:journals/corr/abs-2403-11621}. Recent studies show that subsets of neurons are activated by specific tasks, playing a critical role in LLMs' ability to process knowledge and solve various tasks \citep{DBLP:conf/emnlp/YuA24,DBLP:conf/emnlp/WangCWSL024,DBLP:conf/acl/TangLH0WZWW24}.

\paragraph{Motivation}
Inspired by these findings, we hypothesize that the effectiveness of instruction data may lie in its ability to activate task-relevant neurons. Specifically, when an LLM processes a given sample, the closer the neuron activation pattern aligns with the activation characteristics of the target capability, the more effective the sample is in improving the LLM's performance in that domain. 

\paragraph{Comparison}
Compared to existing methods, the proposed \textsc{Nait} approach offers significant advantages, as shown in Table \ref{tab:methods-comparison}. Specifically, \textsc{Nait} directly leverages neuron activation patterns within LLMs, eliminating the need for external models or complex proxy features and greatly improving data selection efficiency. Furthermore, due to the strong correlation between neuron activation and the internal knowledge and tasks of LLMs, \textsc{Nait} can effectively improve both the model's specialized and general domain capabilities. By capturing the internal neuron activation states during LLM operation, \textsc{Nait} also provides robust interpretability for data selection. Overall, \textsc{Nait} outperforms existing methods in efficiency, scalability, and interpretability, offering a more effective solution for IT data selection.

\label{sec:method}

% The key innovation of \textsc{Nait} lies in establishing quantitative correlations between intermediate activation patterns and the learning potential specific to capability $\mathcal{A}$.

\subsection{Our Data Selection Framework}

% To enhance a specific capability $\mathcal{C}$ in a LLM $\mathcal{M}$, we propose \textsc{Nait}, a novel data selection framework that identifies IT samples that are most effective for improving capability $\mathcal{C}$. Our approach is grounded in the hypothesis that training data which induce $\mathcal{C}$-characteristic activation trajectories in the hidden layers of $\mathcal{M}$ inherently offer greater utility for enhancing the target capability. Below, we demonstrate our proposed method in detail as follows:

\textsc{Nait} identifies IT data that effectively enhances specific domain capabilities $\mathcal{C}$ in the LLM $\mathcal{M}$ via two main modules: (a) Neuron Activation Feature Extraction and (b) Activation Feature-Guided Data Selection, as illustrated in Figure \ref{fig:overall_framework}. The proposed framework is detailed below (the detailed algorithm is provided in Appendix~\ref{app:alg_fra}):

\subsubsection{Module (a). Neuron Activation Feature Extraction for Target Capabilities}
\label{sec:feature_extraction}

To establish the neuron activation features associated with target capabilities $\mathcal{C}$, we implement a two-stage features extraction process:

\begin{itemize}[label=\textbullet, topsep=1pt, itemsep=1pt, left=1pt]
    \item \textbf{In-domain Dataset and Activation Capture}

    We aim to analyze the intrinsic representations of capabilities $\mathcal{C}$ by constructing a representative dataset $\mathcal{P} = \{P_i\}$. Given an in-domain sample $P_i = (t_1, \dots, t_K)$ from $\mathcal{P}$, we record the activations in model $\mathcal{M}$ across the decoder layers $\mathcal{L}$. For a specific layer $l$ and token $t_k$, the activation vector is defined as:
    \begin{equation}
    \mathcal{A}(t_k) = [a_{j}^{(k)}]_{j=1}^J
    \end{equation}
    where $J$ denotes the number of neurons in the layer. Therefore, its relative change $\Delta\mathcal{A}(t_k)$ with respect to the beginning token $b$ can defined as: 
    \begin{equation}
    \Delta\mathcal{A}_i^{(l)} = \mathcal{A}^{(l)}(t_K) - \mathcal{A}^{(l)}(t_1),
    \end{equation}
    where $\Delta\mathcal{A}(t_k) \in \mathbb{R}^J$ quantifies the dynamic activation shifts for each neuron throughout the generation process. 
    To summarize the activations for the entire sequence, we compute the mean activation across all $K$ tokens.
    % :
    %     \vspace{-1em}
    %     \begin{align}
    %     \Delta\mathcal{A}(P_i) &= \frac{1}{K} \sum_{k=1}^K \mathcal{A}(t_k) \in \mathbb{R}^J.
    %     \end{align}
    %     \vspace{-1em}
        
    %     Here, $\Delta\mathcal{A}(P_i)$ represents the average activation state of all $J$ neurons over the sequence $P_i$.
    \item \textbf{Neuron Activation  Direction Extraction} 

        To obtain the characteristic direction vectors $\mathcal{V}$, we apply Principal Component Analysis (PCA) to the difference set $\Delta \mathcal{A}^{(l)}$. For each layer $l$, we extract the first principal component:
    \begin{equation}
        \mathbf{v}_l = \text{PCA}(\Delta \mathcal{A}^{(l)}).
    \end{equation}
    
    To ensure the direction $\mathbf{v}_l$ aligns with the capability's activation trend, we compute the mean difference $\mu_{\text{diff}} = \frac{1}{|\mathcal{P}|} \sum (\mathcal{A}^{(l)}(t_K) - \mathcal{A}^{(l)}(t_1))$. We calibrate the direction as follows:
    \begin{equation}
        \mathbf{v}_l \gets \begin{cases} 
        -\mathbf{v}_l & \text{if } \mu_{\text{diff}} \cdot \mathbf{v}_l < 0 \\
        \mathbf{v}_l & \text{otherwise}
        \end{cases}
    \end{equation}
    The final set of direction vectors is $\mathcal{V} = \{ \mathbf{v}_l \}_{l=1}^L$.
\end{itemize}

\subsubsection{Module (b). Activation Feature-Guided Data Selection}
\label{sec:selection}

Given the instruction dataset $\mathcal{D}_{\text{ins}}$, we perform data selection based on the alignment with the extracted directions $\mathcal{V}$.

\begin{itemize}[label=\textbullet, topsep=1pt, itemsep=1pt, left=1pt]
    \item \textbf{Activation Feature-guided Data Scoring} 
    
    For each sample $y \in \mathcal{D}_{\text{ins}}$, we calculate a score $s_y$ by projecting the sample's activation onto the target direction $\mathbf{v}_l$ and summing across all layers $l=1, \dots, L$:
    \begin{equation}
        s_y = \sum_{l=1}^{L} (\mathcal{A}^{(l)} \cdot \mathbf{v}_l)
    \end{equation}
    This score $s_y$ measures how strongly the sample $y$ activates the specific neurons associated with the target capability.

    \item \textbf{Data Ranking \& Selection} 
    
    Finally, we select the subset $\mathcal{D}_{\text{selected}}$ containing the top-$k$ samples with the highest scores:
    \begin{equation}
        \mathcal{D}_{\text{selected}} = \text{top-}k(\mathcal{S})
    \end{equation}
\end{itemize}

% \paragraph{Theoretical Foundation} 
% The methodology builds on two neuro-symbolic principles:
% \begin{itemize}
%     \item \textbf{Activation Dynamics Principle:} $\Delta\mathbf{AS}$ vectors encode how inputs modify network functionality
%     \item \textbf{Manifold Hypothesis:} PCA projects high-dimensional activations to intrinsic capability subspace
% \end{itemize}
% This combines geometric efficiency (O(d) comparisons vs O(J)) with semantic interpretability.

\section{Experiments}
\label{sec:experiment}

\subsection{Setups}
\label{sec:setups}
% \paragraph{Benchmark}
% We evaluated our approach on various downstream tasks using their data as in-domain data (detailed dataset descriptions are in Appendix \ref{sec:detail_of_train_eval_setup}):
% \begin{itemize}[label=\textbullet, topsep=1pt, itemsep=1pt, left=1pt]
%     \item \textbf{\textit{Factual Knowledge}}: We evaluate the Massive Multitask Language Understanding (MMLU) dataset \citep{DBLP:conf/iclr/HendrycksBBZMSS21}, splitting it into in-domain and test sets, and report 5-shot results.
%     \item \textbf{\textit{Math Capability}}: We use the Grade School Math (GSM) dataset \citep{DBLP:journals/corr/abs-2110-14168}, divided into in-domain and test sets, and report 5-shot results under the CoT setting.
%     \item \textbf{\textit{Reasoning}}: We assess the Big-Bench-Hard (BBH) dataset \citep{DBLP:conf/acl/SuzgunSSGTCCLCZ23}, with in-domain and test splits, and report 5-shot results under the CoT setting.
%     \item \textbf{\textit{Multilingual Understanding}}: We evaluate the TyDiQA benchmark \citep{DBLP:journals/tacl/ClarkPNCGCK20}, a multilingual QA dataset, using in-domain and test splits, and report 5-shot results with the gold-passage setup.
%     \item \textbf{\textit{Coding Capability}}: We evaluate coding ability using BigCodeBench \citep{DBLP:journals/corr/abs-2406-15877} and HumanEval \citep{DBLP:journals/corr/abs-2107-03374}, reporting pass@10 results with a sampling temperature of \texttt{0.8}.
% \end{itemize}
\paragraph{Benchmark}
We evaluate NAIT across five representative capability domains using widely adopted benchmarks: factual knowledge with MMLU \citep{DBLP:conf/iclr/HendrycksBBZMSS21} and MMLU-Pro \citep{DBLP:conf/nips/WangMZNCGRAHJLK24}, mathematical reasoning with GSM \citep{DBLP:journals/corr/abs-2110-14168} and SVAMP \citep{patel-etal-2021-nlp}, general reasoning with BBH \citep{DBLP:conf/acl/SuzgunSSGTCCLCZ23}, multilingual understanding with TyDiQA \citep{DBLP:journals/tacl/ClarkPNCGCK20} and XQuAD \citep{Artetxe:etal:2019}, and coding ability with BigCodeBench \citep{DBLP:journals/corr/abs-2406-15877}, HumanEval (H-Eval) \citep{DBLP:journals/corr/abs-2107-03374} and MBPP \citep{austin2021program}. We use in-domain splits as capability references and conduct evaluation on the corresponding test sets. Benchmark setups are provided in Appendix~\ref{sec:detail_of_train_eval_setup}.
\paragraph{Baselines}
We compare NAIT against a diverse set of representative IT data construction and selection approaches: Alpaca-GPT4 \citep{DBLP:journals/corr/abs-2304-03277}, a widely used self-instruct dataset synthesized by GPT‑4; LIMA \citep{DBLP:conf/nips/ZhouLX0SMMEYYZG23}, which shows that a small amount of carefully curated high-quality IT data can be highly effective; AlpaGasus \citep{DBLP:conf/iclr/ChenLYWGYTS0HJ24}, which leverages ChatGPT to score and filter data; Q2Q \citep{DBLP:conf/naacl/LiZLCC0W0024}, which evaluates data quality via loss signals from a precursor model; and SelectIT \citep{DBLP:conf/nips/Liu0W0WH024}, which selects high-quality data by exploiting uncertainty estimates from base LLMs. Detailed descriptions are provided in Appendix~\ref{sec:detail_of_train_eval_setup}. Additionally, we compare our method with the targeted ability activation approaches, including embedding-based methods, representation-based methods\citep{DBLP:conf/cvpr/ZhangIESW18,DBLP:conf/iclr/HanawaY0I21}, and LESS \citep{DBLP:conf/icml/XiaMGA024}, as detailed in Appendix~\ref{app:alg_fra}.

% \paragraph{Baselines} 
% Our comparison baselines cover a variety of advanced methods for evaluating and improving IT datasets:
% \begin{itemize}[label=\textbullet, topsep=1pt, itemsep=1pt, left=1pt]
%     \item \textbf{\textit{Alpaca-GPT4}}~\citep{DBLP:journals/corr/abs-2304-03277}: A widely utilized IT dataset that leverages a self-instruct methodology, enabling the autonomous generation of instructions through the advanced capabilities of GPT-4.
%     \item \textbf{\textit{LIMA}}~\citep{DBLP:conf/nips/ZhouLX0SMMEYYZG23}: It primarily comprises 1,000 meticulously crafted, high-quality IT data points designed to enhance the alignment capabilities of LLMs. 
%     % \item \textbf{\textit{LESS}\citep{}}: 
%     \item \textbf{\textit{AlpaGasus}}~\citep{DBLP:conf/iclr/ChenLYWGYTS0HJ24}: This approach employs the advanced capabilities of ChatGPT to evaluate and selectively curate data from the original Alpaca dataset.
%     \item \textbf{\textit{Q2Q}}~\citep{DBLP:conf/naacl/LiZLCC0W0024}: It functions by training a precursor model and assessing the quality of the instructional data based on two distinct loss values derived from this model. 
%     % \item \textbf{\textit{Instruction Mining}\citep{}}: This approach involves analyzing data features and loss values to develop a formula for evaluating data quality.
%     \item \textbf{\textit{SelectIT}}~\citep{DBLP:conf/nips/Liu0W0WH024}: A novel method that leverages the intrinsic uncertainty of LLMs to select high-quality IT data without requiring external resources.
    
% \end{itemize}

\renewcommand{\arraystretch}{1.2}
\begin{table*}[t]
\caption{\textbf{Performance comparison of baslines using 10\% of the IT data.} This table shows the results of various tasks across multiple baselines. \textsc{Nait} (e.g., \textsc{Nait} (MMLU)) refers to the process where in-domain dataset to guide the IT data selection. Random refers to a randomly sampled 10\% subset of the IT data. The \textbf{Bold} and \underline{Underline} represent the best and second performance respectively in each column. \textit{$\Delta$ ($\uparrow$)} indicates the performance improvement relative to the ID 01.}
\vspace{1em}
\centering
\scalebox{0.55}{ % 缩放比例为 0.75
\begin{tabular}{cl cc cc cc cc c c c}
\toprule
\textbf{System ID$\downarrow$} & \textbf{Method$\downarrow$} & \multicolumn{2}{c}{\makecell{Factual\\Knowledge}} & \multicolumn{2}{c}{\makecell{Mathematical\\Reasoning}} & \multicolumn{2}{c}{\makecell{Coding\\Ability}} & \multicolumn{2}{c}{\makecell{Multilingual\\Understanding}} & \multicolumn{1}{c}{\makecell{General\\Reasoning}} \\
 & \textbf{Test$\rightarrow$} & \textbf{MMLU} & \textbf{MMLU-Pro} & \textbf{GSM} & \textbf{SVAMP} & \textbf{H-Eval} & \textbf{MBPP}  & \textbf{TydiQA} & \textbf{XQuAD}  & \textbf{BBH} & \textbf{AVG} & \textit{$\Delta$ ($\uparrow$)} \\
\midrule
\multicolumn{13}{c}{\emph{Full Fine-tuning}} \\
\hdashline
01 & Alpaca-GPT4~\citep{DBLP:journals/corr/abs-2304-03277} & 46.87 & 21.89 & 14.63 & 39.00 & \underline{27.87} & \textbf{51.58} & 39.48 & 42.99 & 39.94 & 36.03 & -  \\
02 & LIMA~\citep{DBLP:conf/nips/ZhouLX0SMMEYYZG23} & 45.20 & 23.04 & 15.76 & 37.67 & 27.75 & 46.56 & 44.92 & 44.72 & 39.91 & 36.17 & +0.39\%   \\
03 & 01 + AlpaGasus~\citep{DBLP:conf/iclr/ChenLYWGYTS0HJ24} & 43.21 & 21.96 & 13.34 & 36.67 & 23.94 & 46.08 & 44.70 & 46.84 & 39.91 & 35.18 & -2.34\%  \\
04 & 01 + Q2Q~\citep{DBLP:conf/naacl/LiZLCC0W0024} & 46.73 & 21.50 & 14.50 & 35.00 & 25.19 & 44.97 &  44.41 & 48.44 & 40.34 & 35.68 & -0.98\% \\
05 & 01 + SelectIT~\citep{DBLP:conf/nips/Liu0W0WH024} & \bf{47.90} & 22.86 & 15.40 & \underline{41.11} & 27.92 & 49.47 & 43.91 & 45.56 & 40.33 & 37.16 & +3.15\%  \\
06 & 01 + Random Baseline & 47.14 & 21.43 & 14.13 & 35.67 & 25.55 & 47.35 & 44.16 & 46.56 & 39.21 & 35.69 & -0.94\% \\
\hdashline
\multicolumn{13}{c}{\emph{Our Proposed Method (Individual Capability Features)}} \\
\hdashline
07 & 01 + \textsc{Nait} (MMLU) & \underline{47.81} & 23.61 & 15.68 & 39.67 & 25.23 & 47.47 & \underline{47.16} & \textbf{49.47} & 38.52 & 37.18 & +3.20\% \\
08 & 01 + \textsc{Nait} (GSM) & 46.45 & \textbf{24.50} & \underline{16.00} & \textbf{41.33} & 27.84 & 48.41 & 46.54 & 47.97 & 40.28 & \textbf{37.70} & \textbf{+4.65\%} \\
09 & 01 + \textsc{Nait} (CodeX) & 47.51 & \underline{23.46} & 15.53 & 38.67 & \bf{28.49} & \underline{49.74} & 44.19 & 46.27 & 39.72 & 37.06 & +2.88\%  \\
10 & 01 + \textsc{Nait} (TydiQA) & 46.17 & 22.82 & 13.80 & 37.67 & 25.02 & 47.88 & \bf{47.78} & \underline{49.23} & 40.00 & 36.71 & +1.89\%   \\
11 & 01 + \textsc{Nait} (BBH) & 47.78 & 23.36 & 13.34 & 36.67 & 25.15 & 47.08 & 45.93 & 48.46 & \bf{40.46} & 36.47 & +1.23\%  \\
\hdashline
\multicolumn{13}{c}{\emph{Our Proposed Method (All Capability Features)}} \\
\hdashline
12 & 01 + \textsc{Nait} (7 - 11) & 46.83 & 23.29 & \bf{16.53} & 39.67 & 26.44 & 47.62 & 46.09 & 48.27 & \underline{40.02} & \underline{37.20} & \underline{+3.24\%} \\
\bottomrule
\end{tabular}
}
\vspace{-5mm}
\label{table:main LLMs}
\end{table*}

\paragraph{Implementation Details}
In this study, we adopt LLaMA-2-7b as the foundational model for fine-tuning. The detailed fine-tuning and text generation settings are provided in Appendix~\ref{sec:detail_of_train_eval_setup}. 
% \bf{mention use VLLM for evaluating}

\subsection{Main Results}
\label{sec:main_result}
% \textsc{Nait} 方法通过利用与多种特定能力相关的神经元激活特征，有效提升了模型的整体性能。如图 \ref{table:main LLMs} 所示，ID 01 代表原始 Alpaca-GPT4 的基准性能，而 ID 12 则展示了所提出的 \textsc{Nait} 方法在完全监督的 IT 环境下超越了 Alpaca-GPT4 值得注意的是，该方法仅使用了原始数据的 10%，却在五个测试数据集上实现了高达 3.00% 的性能提升。这一结果不仅显著优于现有的 IT 数据选择方法，如 AlpaGasus（ID 03）、Q2Q（ID 04）和 SelectIT（ID 05），而且进一步证明了 \textsc{Nait} 方法在数据利用效率和泛化能力方面的优势。然而，我们也观察到，不同的激活能力之间可能存在相互增强或相互抑制的作用，例如ID 10在单独使用TydiQA激活其域内能力时，数学和代码的表现出现了退化现象，然而在ID12当激活期域内能力时，数学的能力又得到了增强种相互作用在一定程度上会影响模型在其他下游任务中的表现。

\paragraph{Overall Performance}
The \textsc{Nait} method effectively enhances model's overall performance by leveraging neuron activation features across multiple domain capabilities. As shown in Table \ref{table:main LLMs}, System 01 represents the baseline performance of the Alpaca-GPT4, while System 12 demonstrates the proposed \textsc{Nait} method surpassing Alpaca-GPT4 under full fine-tuning setting. Notably, it achieves up to a \textbf{4.65\%} improvement using mathematical reasoning features (System 08) and a \textbf{3.24\%} improvement when combining all capability features (System 12). These results significantly surpass existing IT data selection methods, such as AlpaGasus (System 03), Q2Q (System 04), and SelectIT (System 05), while utilizing only 10\% of the original data. We also observe that activating specific domain capabilities influences downstream tasks differently. For instance, utilizing only the multilingual capability (System 10) leads to decreased performance on mathematical and coding tasks compared to the baseline. In contrast, when all in-domain capabilities are activated (System 12), the model's mathematical performance notably improves from 14.64 to 16.53, demonstrating the robustness of \textsc{Nait}.

% ID 07 至 11 的实验结果展示了 \textsc{Nait} 方法在针对个体特定能力时，能够实现更为一致且稳定的性能提升。值得注意的是，这一优势在诸如 GSM、TydiQA 和 CodeX 等复杂任务中表现得尤为显著，分别较随机基线提升了 1.87、3.62 和 2.94 个百分点。进一步分析表明，神经元激活特征在不同任务领域中的贡献存在差异，并表现出一定的跨领域迁移能力。例如，基于 BBH 数据集提取的神经元激活特征能够显著提升模型在 MMLU 任务上的表现，而基于 CodeX 数据集提取的神经元激活特征同样能够有效提升模型在 GSM 任务上的性能，且提升幅度均接近了各自原始激活状态下的表现。值得一提的是，\textsc{Nait} 方法在 GSM 数据集上的性能提升高达 3.80%，甚至超越了多能力共同激活下的模型。这一结果充分说明，\textsc{Nait} 方法不仅能够有效剔除冗余或高度相似的样本，还能在保持数据多样性的同时，进一步验证其在优化数据利用和提升特定任务性能方面的有效性与泛化能力。

% 模型在大多数任务中的整体表现达到了35.42\%，唯独在 MMLU 任务上未能带来预期的增益仅为46.56。对于这一现象的具体原因及其影响，将在第 \ref{sec：task_transferability} 节中进行深入讨论。值得一提的是，\textsc{Nait} 在 GSM 数据集上的表现尤为突出，性能提升高达 3.80%，甚至超越了多能力共同激活下的模型。这一结果充分说明 \textsc{Nait} 方法不仅能够有效剔除冗余或高度相似的样本，同时还能保持数据的多样性，进一步验证了其在优化数据利用、提升特定任务性能方面的有效性和泛化能力。

\paragraph{\textsc{Nait} with Distinct Capability Features}
The experimental results from System 07 to 11 demonstrate that the \textsc{Nait} method can achieve more consistent and stable performance improvements when targeting individual-specific abilities. Notably, this advantage is particularly pronounced in complex tasks such as mathematical reasoning, multilingual understanding, and coding ability, with improvements of 7.53\% and 6.29\%, and 5.33 \% over the random baseline, respectively. Further analysis reveals that the contributions of neuron activation features vary across different task domains and exhibit a certain degree of cross-domain transferability. For instance, neuron activation features extracted based on the MMLU in-domain dataset can significantly enhance the model’s performance on the multilingual understanding task, while those extracted from the CodeX in-domain dataset can likewise effectively improve the model’s performance on the GSM task. It is worth mentioning that the performance of the \textsc{Nait} method on the GSM dataset reaches as high as 4.65\%, even surpassing the model under multi-capability activation (System 12). Additionally, we compare our method with targeted ability activation approaches, including embedding-based methods, representation-based methods \citep{DBLP:conf/cvpr/ZhangIESW18,DBLP:conf/iclr/HanawaY0I21}, and LESS \citep{DBLP:conf/icml/XiaMGA024}, as detailed in Appendix \ref{app:nait_vs_less}. Overall, \textsc{Nait} consistently achieves the superior overall average performance across all targeted settings. While gradient-based methods like LESS can outperform in the specific target domain to some extent, this specialization often comes at the cost of generalization, leading to performance degradation in non-target tasks. In contrast, \textsc{Nait} demonstrates robust transferability, maintaining high proficiency across multilingual and reasoning tasks regardless of the target feature.

\section{Analysis}
\label{sec:analysis}
This section aims to analyze further and address the following research questions: (1) How do the number of in-domain datasets, the proportion of IT datasets, model size, and IT data methods affect \textsc{Nait}? (see \S~\ref{sec:ablation}) (2) What are the advantages of \textsc{Nait} in cost efficiency? (see \S~\ref{sec:cost_efficiency}) (3) How interpretable in \textsc{Nait}? (see \S~\ref{sec:interpretability}) (4) How does \textsc{Nait} perform in target domain? (\S~Appendix~\ref{sec:IT_for_targeted_domain})

\subsection{Ablation Study}
\label{sec:ablation}
\begin{wrapfigure}{r}{.5\textwidth}
    \vspace{-2em}
    \centering
    \includegraphics[width=0.48\textwidth, trim=0 0 0 0]{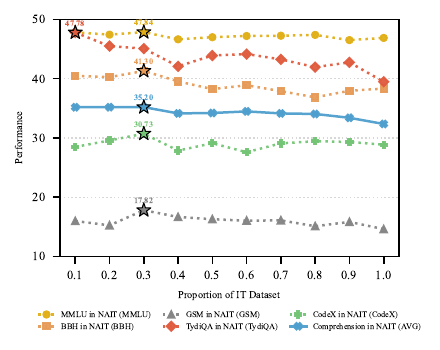}
    \vspace{-5mm}
    \caption{\textbf{Performance at different proportions of the IT dataset.} Task performance in \textsc{Nait} (the in-domain dataset), e.g., MMLU in \textsc{Nait}, refers to using in-domain data to guide IT data selection and to assess task outcomes. The comprehensive results correspond to System 12.}
    \vspace{-1em}
    \label{fig:ab_ratio}
\end{wrapfigure}

\paragraph{Proportion of Selected IT Dataset}
% 尽管 \textsc{Nait} 在样本评估与排序方面展现出优异的性能，如何从包含大量冗余信息的数据集中选取最优的数据比例，依然是当前方法亟需解决的关键问题。为进一步探究该问题，我们通过系统性地调整微调训练集的规模（范围为 10% 至 100%），分析了 \textsc{Nait} 在不同数据比例下的性能变化。如图 \ref{fig:ab_ratio} 所示，\textsc{Nait} 筛选出的前 30% IT 样本表现尤为突出，仅使用前 30% 样本进行训练时，LLM 达到了最佳性能。值得注意的是，随着训练数据比例的持续增加，模型性能不仅出现了更大波动，且整体呈下降趋势，在使用全部数据（100%）时降至最低。上述结果表明，过多的冗余数据可能削弱模型的泛化能力，进一步强调了合理筛选训练数据对于提升模型性能的重要性。
Although \textsc{Nait} demonstrates excellent performance in data evaluation and ranking, selecting the optimal data proportion from IT datasets with a large amount of redundant information remains challenging. To further investigate this, we adjusted the proportion of the selected IT dataset from 10\% to 100\% and analyzed the performance of \textsc{Nait} under different IT data proportions. As shown in Figure~\ref{fig:ab_ratio}, the top 30\% of IT samples selected by \textsc{Nait} exhibited the best performance. Notably, as the proportion of training data continued to increase, the model’s performance showed an overall downward trend, reaching its lowest point when the entire dataset (100\%) was used. These results indicate that excessive redundant data may undermine the model’s generalization, further emphasizing the need for data selection to maximize model performance.

\begin{figure}[!h]
    % \vspace{1em}
    \centering
    \includegraphics[width=0.9\textwidth, trim=0 20 0 20]{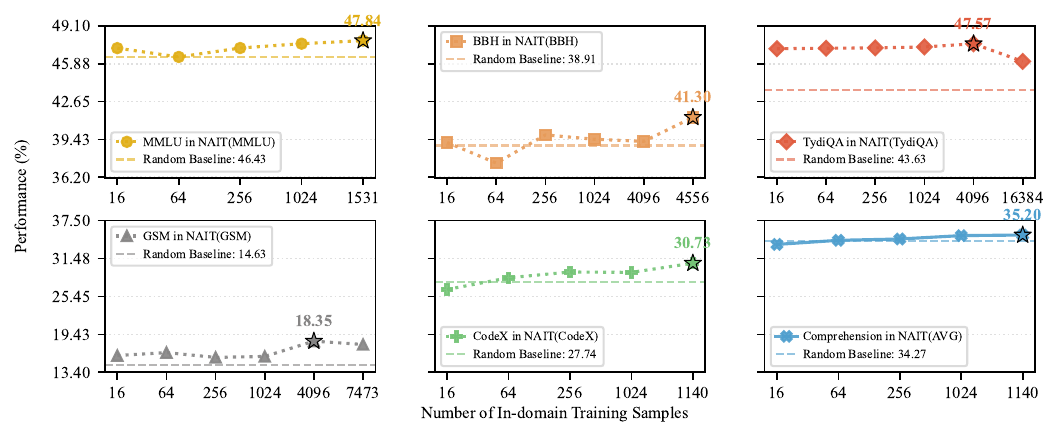}
    \caption{\textbf{Comparison of \textsc{Nait} with random data selection across different in-domain data scales at 30\% IT dataset.} Task performance in \textsc{Nait} (in-domain dataset), such as MMLU in \textsc{Nait} (MMLU), refers to using the in-domain dataset to guide IT data selection and evaluate task performance. The dashed line indicates the 30\% of the IT dataset random selection baseline.}
    \label{fig:ab_ntrain}
    % \vspace{-1em}
\end{figure}

\paragraph{Number of In-domain Data}
% We evaluated the impact of the sample number of dataset $\mathcal{P}$ used to grasp the neuron activity shown in \autoref{fig:ab_ntrain}.
% 域内数据的数量是本方法中的关键参数，因为它直接影响针对特定模型能力提取神经元激活特征的有效性。如图2所示，我们对每一组域内数据激活下筛选的IT数据进行微调，并评估其在相应任务上的表现。结果表明，\textsc{Nait}方法在所有任务中的性能在绝大多数情况下均优于随机选择基线，尤其在数据量较小（如16、64、256条）时表现尤为显著，表明此时所提取的特征已能较好地捕捉所需的能力。随着域内数据量的增加，性能提升趋于平稳，大多数任务在域内数据量达到最大时取得峰值。然而，GSM和TydiQA任务在数据量为4096时已分别达到18.35和447.57的性能，这表明域内数据的质量同样对\textsc{Nait}方法捕捉所需能力具有重要影响。
The number of in-domain data is a critical parameter in our method, as it directly influences the effectiveness of extracting neuron activation features for specific domain capabilities. As shown in Figure~\ref{fig:ab_ntrain}, we fine-tune the model using IT data selected based on the activation patterns of each in-domain data group and evaluate its performance on corresponding tasks. The results indicate that \textsc{Nait} outperforms the random selection baseline across all tasks in most cases. Even when the data size is small (e.g., 16, 64, or 256 samples), most cases exceed the random baseline. This suggests that even with limited data, the extracted features effectively capture the required capabilities. As the number of in-domain data increases, performance improvements gradually level off, with most tasks reaching their maximum performance at the largest data size. However, for GSM and TydiQA, peak performance is achieved at 18.35 and 47.57, respectively, when the data size is 4096. This highlights that the quality of in-domain data also plays an important role in enabling \textsc{Nait} to capture task-specific features.

\paragraph{Intensity of Neuron Activation}
% % 我们检验\textsc{Nait}的有效性。我们的比较方法包括随机基线，高神经元激活特征和低神经元激活特征。结果强有力地支持了我们方法的有效性，如图x所示，\textsc{Nait} with高神经元激活特征的性能有效地大于随机基线的性能，这说明\textsc{Nait} with高神经元激活特征可以选择的样本可以有效激活模型的指令能力；随机基线的性能大于\textsc{Nait} with低神经元激活特征，这说明\textsc{Nait} with低神经元激活特征所选择的样本抑制了模型的相关的指令能力，或者对模型相关的能力造成了损伤。
\begin{wraptable}{r}{0.5\textwidth} % specify width
\vspace{-2em}
\centering
\large
\caption{\textbf{Performance comparison across datasets at 10\% IT dataset under different neuron activation levels.}}
\vspace{1em}
\scalebox{0.5}{% 缩放比例更小以适应
\begin{tabular}{c c c c c c c c}
\toprule
\multirow{2}{*}{\textbf{Intensity}} & \multirow{2}{*}{\textbf{MMLU}} & \multirow{2}{*}{\textbf{BBH}} & \multirow{2}{*}{\textbf{H-Eval}} & \multirow{2}{*}{\textbf{TydiQA}} & \multirow{2}{*}{\textbf{H-Eval}} & \multicolumn{2}{c}{\textbf{Overall}} \\
\cmidrule(lr){7-8}
& & & & & & \textbf{AVG} & \textit{$\Delta$ ($\uparrow$)} \\
\midrule
Random & \bf{47.14} & 39.21 & 14.13 & 44.16 & 25.55 & 34.04 & - \\
High & 46.83 & \bf{40.02} & \bf{16.53} & \bf{46.09} & \bf{26.44} & \bf{35.18}  & +3.35\% \\
Low & 46.30 & 28.15 & 10.61 & 36.12 & 20.15 & 28.27 & -17.54\% \\
\bottomrule
\end{tabular}%
}
% \vspace{-4mm}
\label{tab:ab_intensity}
\end{wraptable}
We evaluate the effectiveness of \textsc{Nait} by comparing three \textsc{Nait} settings: \textbf{Random} (10\% of samples randomly selected), \textbf{High} (top 10\% with highest neuron activation feature alignment scores), and \textbf{Low} (10\% with lowest neuron activation feature alignment scores). As shown in Table \ref{tab:ab_intensity}. Specifically, \textsc{Nait} with highest neuron activation feature alignment scores outperforms the Random baseline by 3.35\%, indicating that these samples effectively enhance the models' capabilities. In contrast, \textsc{Nait} with lowest neuron activation feature alignment scores reduces performance by 17.54\% compared to Random baseline, suggesting that such samples not only fail to help but also negatively impact the LLM's effectiveness.

\begin{wraptable}{t}{0.45\textwidth} % 参数 t 表示顶部对齐, 宽度占页面的 90%
\vspace{-1em}
\caption{\textbf{Effectiveness of \textsc{Nait} Across Models.} Alpaca and Random are trained on the full dataset and a random of 10\% subset, while \textsc{Nait} selects a 10\% subset.}
\vspace{1em}
\centering
\scalebox{0.55}{ % 调整缩放比例
\begin{tabular}{cccccccc}
\toprule
\multirow{2}{*}{\textbf{Method}} & \multirow{2}{*}{\textbf{MMLU}} & \multirow{2}{*}{\textbf{BBH}} & \multirow{2}{*}{\textbf{GSM}} & \multirow{2}{*}{\textbf{TydiQA}} & \multirow{2}{*}{\textbf{H-Eval}} & \multicolumn{2}{c}{\textbf{Overall}} \\
\cmidrule(lr){7-8}
& & & & & & \textbf{AVG} & \textit{$\Delta$ ($\uparrow$)} \\
\midrule
\multicolumn{8}{c}{\emph{LLama-2-13b}} \\
\hdashline
Alpaca        & 53.90 & 45.00 & 20.85 & 44.13 & 34.84 & 39.74 & - \\
+Random   & 52.70 & \bf{47.59} & 22.06 & 44.66 & 36.79 & 40.76 & +2.57\% \\
+\textsc{Nait} & \bf{54.10} & 47.04 & \bf{24.11} & \bf{48.89} & \bf{38.53} & \bf{42.53} & \bf{+7.02\%} \\
\midrule
\multicolumn{8}{c}{\emph{Mistral-7b}} \\
\hdashline
Alpaca        & 47.00  & 40.19 & 12.89 & 35.40 & 34.42 & 33.98 & - \\
+Random   & 48.37 & 46.11 & 17.36 & 34.80 & 41.81 & 37.69 & +10.92\% \\
+\textsc{Nait}       & \bf{52.90} & \bf{49.07} & \bf{18.95} & \bf{41.22}& \bf{45.02} & \bf{41.43} & \bf{+21.92\%}  \\
\midrule
\multicolumn{8}{c}{\emph{Llama-3-8b}} \\
\hdashline
Alpaca        & 59.60 & 48.38 & 29.11 & 48.38 & 57.36 & 48.57 & - \\
+Random   & 48.37 & 55.93 & 37.23 & \bf{54.65} & 69.63 & 52.10 & +7.27\% \\
+\textsc{Nait}       & \bf{60.80} & \bf{60.93} & \bf{43.91} & 47.72 & \bf{74.78} & \bf{57.63} & \bf{+18.65\%} \\
\midrule
\multicolumn{8}{c}{\emph{Qwen-2.5-7b}} \\
\hdashline
Alpaca   & 73.30 & 65.46 & 83.17 & 49.33 & 90.85 & 72.42 & - \\
+Random   & 74.00 & 67.78 & \textbf{84.31} & 51.25 & 89.63 & 73.39 & +1.34\% \\
+\textsc{Nait} & \textbf{74.21} & \textbf{67.87} & 83.70 & \textbf{58.14} & \textbf{92.07} & \textbf{75.20} & \textbf{+3.83\%} \\

\bottomrule
\end{tabular}}
% \vspace{-3em}
\label{table:various_model}
\end{wraptable}
\paragraph{Various Foundation Models}
Our method demonstrates significant performance improvements when fine-tuning the \emph{LLaMA-2-7b} model with the curated Alpaca-\textsc{Nait} dataset. To evaluate its robustness and generalizability, we extend our study to other foundational models, including (1) \emph{LLaMA-2-13b}, representing models of different scales; and (2) \emph{Mistral-7b} and \emph{LLaMA-3-8b}, representing models with different architectures, and (3) \emph{Qwen-2.5-7b}, representing state-of-the-art models with stronger baselines. As shown in Table \ref{table:various_model}, \textsc{Nait} consistently enhances performance across all evaluated models. Specifically, it maintains effectiveness on larger scales (\emph{LLaMA-2-13b}) and achieves substantial gains of +21.92\% on \emph{Mistral-7b} and +18.65\% on \emph{LLaMA-3-8b}. Notably, even on the highly capable \emph{Qwen-2.5-7b}, \textsc{Nait} further boosts the average performance by +3.83\%, outperforming the random baseline and validating its efficacy regardless of the model's initial capabilities.

\paragraph{Different Instruction Tuning Datasets}

\begin{wrapfigure}{r}{.45\textwidth}
    \vspace{-2em}
    \centering
    \includegraphics[width=0.46\textwidth, trim=0 0 0 0]{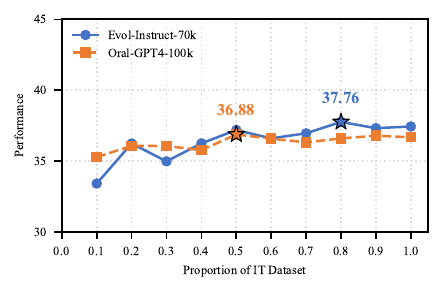}
    \vspace{-2em}
    \caption{\textbf{Comparison of \textsc{Nait} across different IT dataset.} Task performance in \textsc{Nait} refers to using the all in-domain dataset to guide IT data selection and evaluate task performance. }
    \label{fig:rob_IT_dataset}
    \vspace{-3mm}
\end{wrapfigure}
% 将这种方法的稳健性扩展到其他指令数据集也同样重要。我们进一步验证我们方法的有效性在 WizardLM 的 Evo-Instruct IT数据集上和Orcal。结果可以在表格 \autoref{table:various_instruction_dataset} 中显示，
To further verify the effectiveness of our method, we extend our IT selection method to other IT datasets, including Evo-Instruct~\citep{DBLP:conf/iclr/XuSZG0FTLJ24} using WizardLM and Orca-GPT4~\citep{DBLP:journals/corr/abs-2306-02707}. The experimental results are shown in Figure~\ref{fig:rob_IT_dataset}, as the proportion of IT data increases, the model performance rises and then declines. Notably, the efficient IT data subsets selected by \textsc{Nait} achieve optimal performance at 50\% (Orca-GPT4) and 80\% (Evo-Instruct), significantly outperforming the use of the full dataset. Additionally, Evol-Instruct and Orca-GPT4 require a higher proportion of data, especially Evol-Instruct, which needs up to 80\% to achieve optimal performance, reflecting their higher quality, complexity, and information density compared to Alpaca-GPT4.

\paragraph{Different Selection Strategies} We also conducted experiments using context length and perplexity as measures of data complexity, following \citet{DBLP:conf/iclr/0131Z00H24}, to provide an additional insight into our method's effectiveness, as detailed in Appendix~\ref{sec:different_selection_strategy}.

% \subsection{Insights of Selective Data Curation}
% \cx{provide each selection data example}

% \paragraph{Neuron Activity Similarity Analysis}

% \paragraph{Data Characteristic Analysis}

\subsection{Cost Efficiency}
\label{sec:cost_efficiency}
\begin{wraptable}{r}{0.5\textwidth} % r=右侧放置, 宽度设为55%正文宽度
\vspace{-2em}
\caption{\textbf{Efficiency Comparison of Different Methods on NVIDIA A800 80GB with Batch Size set to 8}. API costs follow the official OpenAI pricing (\$2.00/million input tokens, \$8/million output tokens for GPT-4.1); GPU costs are estimated based on the Google Cloud pricing (\$1.15 per GPU hour for NVIDIA A800 80GB).}
\vspace{1em}
\centering
\scalebox{0.6}{ % 缩放比例，可调大或调小
\begin{tabular}{ccrr}
\toprule
{\textbf{Method}} & {\textbf{Externally-Independent}} & {\textbf{Time}} & {\textbf{Cost}} \\
\midrule
AlpaGasus\citep{DBLP:conf/iclr/ChenLYWGYTS0HJ24} & \ding{55} & 19.07h & \$178.02 \\
Q2Q\citep{DBLP:conf/naacl/LiZLCC0W0024} & \ding{55} & 3.52h & \$4.05 \\
LESS\citep{DBLP:conf/nips/Liu0W0WH024} & \ding{51} & 9.86h & \$11.33 \\
SelectIT\citep{DBLP:conf/icml/XiaMGA024} & \ding{51} & 23.20h & \$26.68 \\
\textsc{Nait} & \ding{51} & 1.32h & \$1.52 \\
\bottomrule 
\end{tabular}
}
\vspace{-5mm}
\label{tab:Efficiency}
\end{wraptable}
% 本研究对比分析了所提出方法与 AlpaGasus（GPT-4o）、Q2Q（GPT-4o）以及 SelectIT 方法在 Alapa-GPT4 数据集上的模型推理时间和费用成本表现。表1的结果表明，与外部依赖模型相比，\textsc{Nait} 方法分别比 AlpaGasus 和 Q2Q 节省了约19倍和4倍，推理速度分别提升了 XX 和 XX 秒。此外，相较于外部独立方法 SelectIT，我们的方法在费用成本上降低了 XX%，优化后的架构实现了 X.X 倍的速度提升。上述结果验证了在现有外部独立方法基础上进行高效资源整合的可行性和有效性。
We compare inference time and cost between \textsc{Nait} and existing approaches, including AlpaGasus (GPT-4o), Q2Q (GPT-4o), and SelectIT, on the Alpaca-GPT4 dataset (52k). As shown in Table~\ref{tab:Efficiency}, \textsc{Nait} achieves up to 19× and 4× cost reductions compared to AlpaGasus and Q2Q, respectively, along with significant improvements in inference speed by 17.75 and 2.2 hours. Compared to SelectIT, \textsc{Nait} reduces cost by 94.3\% and achieves a 17.58× speedup. These results highlight the feasibility and effectiveness of \textsc{Nait} in real-world applications.
% In this study, we conducted a comparative analysis of our proposed method and existing approaches, including AlpaGasus (GPT-4o), Q2Q (GPT-4o), and SelectIT, on the Alapa-GPT4 dataset (52k), focusing on model inference time and cost. Compared to external dependency models, the results in Table \ref{tab:Efficiency} demonstrate that the \textsc{Nait} method reduces costs by about 19 times and 4 times savings relative to AlpaGasus and Q2Q, respectively, while also accelerating processing speed by 17.75 and 2.2 hours. Furthermore, when compared with the external independence method SelectIT, our approach achieves a cost reduction of 94.30\% and an optimized architecture that delivers a 17.58× speedup. These findings validate the feasibility and effectiveness of efficient resource integration based on existing external-independence methods.

\subsection{Interpretability Analysis}
\label{sec:interpretability}
\paragraph{Transferability of Neural Activation Feature}
From the results in Table~\ref{table:main LLMs}, we observe that the neural activation features corresponding to different capabilities exhibit remarkable variation in their transferability. For instance, the neural activation feature extracted based on GSM significantly enhances the overall average performance of LLMs, and it also yields positive effects on BBH and CodeX tasks. To further investigate the relationships between the transferability of neural activation features, we introduce \textbf{Transferability} to evaluate each capability during the transfer process. Transferability of capability $i$ reflects to what extent $i$ can enhance other capabilities when activated. It is defined as
\begin{equation}
   \text{Transferability}_i =  \text{Acc}(i, j) - \text{Acc}(j, j)
\end{equation}
where $\text{Acc}(i, j)$ is the performance on task $j$ activated by feature $i$, and $\text{Acc}(j, j)$ is task $j$’s baseline performance with its own activation feature. 
\begin{figure*}[!h]
    \centering
    \vspace{-1mm}
    \includegraphics[width=1.0\textwidth, trim=0 30 0 10]{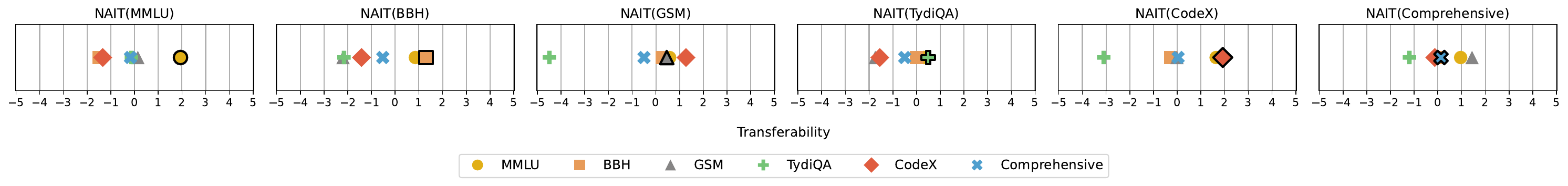}
    \caption{\textbf{Transferability of neural activation features across capabilities.} Each column represents the capability to which the neural activation feature is applied. Icons with an outline border indicate the capability’s performance on its own task, which serves as the baseline reference.}
    \vspace{-2mm}
    \label{fig:ia_transferability}
\end{figure*}

\begin{wrapfigure}{r}{.4\textwidth}
    \centering
    \vspace{2em}
    \includegraphics[width=0.4\textwidth, trim=0 0 0 0]{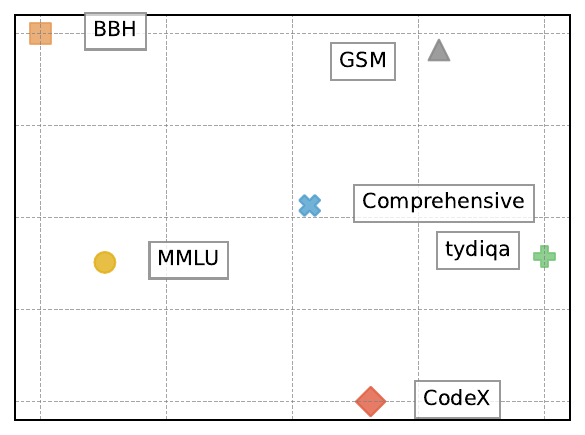}
    \caption{\textbf{Comparison of the direction of activation feature extracted by \textsc{Nait} across different capabilities.}}
    \label{fig:ia_direction}
    \vspace{-3em}
\end{wrapfigure}

As shown in Figure~\ref{fig:ia_transferability}, the transferability across capabilities can be directly observed. On the one hand, activation features of single capabilities such as GSM and CodeX exhibit evident positive transfer in cross-task settings, indicating that logical reasoning and programmatic features possess strong general power. In contrast, the activation features of TydiQA show weaker or even negative transfer, and their performance within the task itself is relatively limited, reflecting both a dependence on language-specific patterns and shortcomings in cross-domain adaptability. On the other hand, when multiple capabilities are aggregated into a comprehensive activation direction, the model achieves optimal overall transferability. This finding demonstrates the existence of a universal core feature whose neural activation patterns remain stable across tasks, thereby providing a solid foundation for the model’s integrated capabilities. 
\paragraph{Direction of Activation Features}
\label{sec:task_direction}
To further investigate the interpretability of NAIT, we apply T-SNE to project the neuron activation features into representation space (see~\autoref{fig:ia_direction}). The figure shows that the feature's directions corresponding to different targeted capabilities form relatively separable clusters in the representation space, indicating their consistency with the task-specific reasoning mechanisms or knowledge domains. Notably, the comprehensive, as an aggregated representation across multiple tasks, reveals the \textbf{general core capabilities.} This indicates that there exist general data or features capable of simultaneously activating multiple distinct abilities, thereby supporting cross-task knowledge transfer and capability integration. 
\paragraph{Selected IT Data Distribution}
To further analyze the distribution of IT subsets selected by different activation features, we performed both qualitative and quantitative analyses (see Appendix \ref{app:case_study}). Figure \ref{fig:ia_data_distribution} shows the distribution of the 10\% IT dataset selected by \textsc{Nait}. The results reveal that 58.87\% of the subsets overlap across different capabilities, indicating that \textsc{Nait} consistently identifies a stable set of \textbf{general core data} applicable across tasks. However, GSM8K exhibits the highest demand for task-specific data, with 1,034 unique samples, while other tasks like MMLU and CodeX rely more heavily on shared core data. This demonstrates \textsc{Nait}'s ability to balance general-purpose data selection with the retention of task-specific samples, ensuring robust performance across diverse benchmarks. Qualitatively, unlike embedding-based methods that focus on surface-level similarity, \textsc{Nait} prioritizes samples requiring complex task-following and logical reasoning. While all methods capture the fundamental formats associated with factual knowledge, they consistently select the corresponding factual multiple-choice questions.
\begin{figure*}[!h]
    \centering
    % \vspace{-1em}
    \includegraphics[width=1.0\textwidth, trim=0 0 0 0]{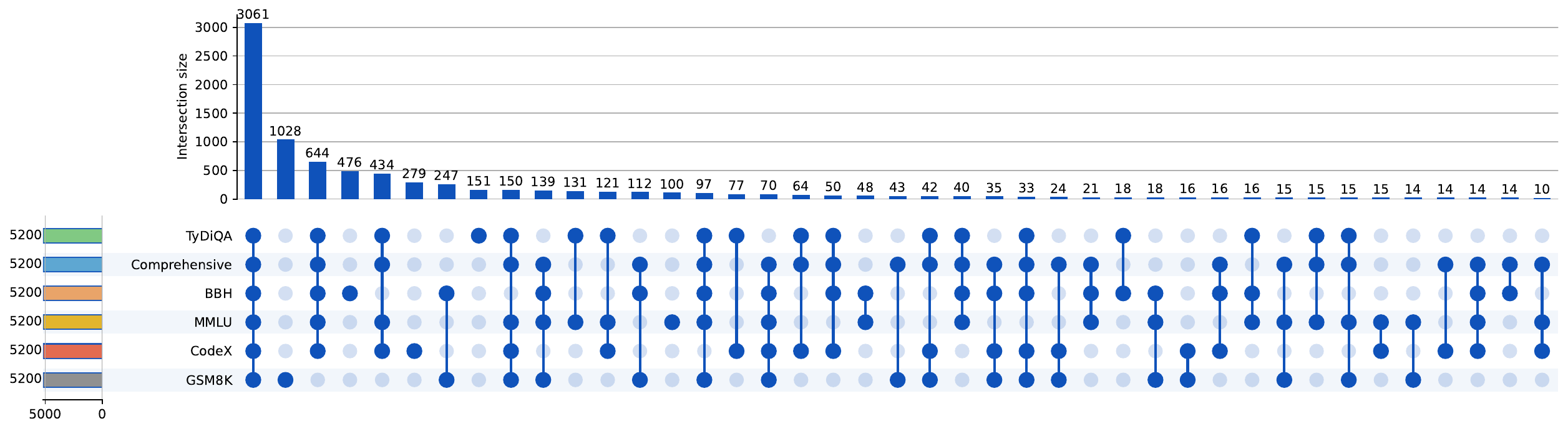}
    % \vspace{-2em}
    \caption{\textbf{Distribution of 10\% IT dataset selected by \textsc{Nait}.} The upper bar chart quantifies the intersection sizes among different IT data subsets, while the connected dots below identify the specific capability combinations corresponding to each intersection.}
    \label{fig:ia_data_distribution}
    \vspace{-2em}
\end{figure*}

% This distribution pattern further validates the presence of a universal subset of data that underlies cross-task representational consistency, serving as the foundation for the transferable activation directions observed in \autoref{fig:ia_transferability} and \autoref{fig:ia_direction}. 

\section{Conclusion}
\label{sec:Conclusion}

% 本文介绍了一种用于 LLM 指令调整的新型数据选择策略 \textsc{Nait}，该策略利用 LLM 的内部神经元的神经元激活特征来高效识别高质量 IT 数据，而无需额外资源。\textsc{Nait} 包括两个主要步骤：（1）Extracting Neuron Activation Features for Capability A；（2）Instruction Data Selection 。通过将 \textsc{Nait} 应用于 Alpaca-GPT4 数据集，我们引入了一个紧凑而强大的 IT 数据集，称为 \textsc{Nait} Alpaca。不同的模型和指令数据集和领域任务上的实验证明了 \textsc{Nait}的有效性。

In this paper, we propose \textsc{Nait}, an efficient framework for selecting high-quality IT data by leveraging neural activation patterns. \textsc{Nait} identifies the optimal data subset by evaluating the alignment between candidate samples and neural activation features associated with target capabilities. Experimental results demonstrate that models fine-tuned on selected data consistently outperform baseline methods. Ablation studies further verify that only a small number of samples are sufficient to accurately extract activation features that capture model capabilities. In addition, \textsc{Nait} exhibits strong robustness and reveals significant cross-task transferability of neural activation features. Further analysis shows that data with logical reasoning and programmatic characteristics tend to activate stronger general capabilities in the model, and that a stable core subset of data exists that can universally enhance performance across a variety of tasks.

% 本文提出\textsc{Nait}，一种高效的框架，通过利用神经激活模式来选择高质量的IT数据。\textsc{Nait}通过评估其与与目标域能力相关的神经激活特征的对齐来确定最佳数据集。实验结果表明，在\textsc{Nait}选择的数据上微调的模型始终优于基线方法，消融实验表明只需要少量样本就可以找到激活模型能力的特征。并且NAIT具有很强的鲁棒性，并展示了神经激活特征跨任务的强迁移性。 
% 此外，进一步分析发现具有逻辑推理和程序设计的数据具备激活模型强大的综合能力。并且存在一个普遍的核心数据让模型在各项能力上都得到genral的加强
% 这些发现强调了多样化和复杂的激活特征在增强llm的通用能力方面的关键作用，为优化IT数据选择提供了一种有效和可扩展的方法。

% \paragraph{IT Data Quantity and IT Datasets} Our findings suggest that prioritizing the top 10\% of high-quality data enhances Alpaca's performance. However, this proportion has only been validated on the Alpaca dataset and may not generalize to other instruction datasets. Future research could explore dynamically determining the optimal proportion of high-quality data based on the quality distribution of different datasets to improve adaptability.

% 不同规模的模型

% 数据集的数量

\section*{Ethics statement}
This work adheres to the ICLR Code of Ethics.\footnote{ \url{https://iclr.cc/public/CodeOfEthics}} Our research focuses on improving the efficiency and interpretability of IT for LLMs through neurally informed data selection. We do not introduce new data collection involving human participants, nor do we use private or personally identifiable data. All datasets employed (e.g., Alpaca-GPT4, GSM8K, MMLU, BBH, TyDiQA, CodeX) are publicly available and widely used in the research community. We carefully respect data licenses and ensure compliance with privacy and intellectual property standards. Our method aims to reduce the computational and environmental cost of developing LLMs by selecting smaller subsets of high-quality data, which aligns with the principle of minimizing harm. However, as with other LLM research, there is the potential risk of misuse in generating biased or harmful outputs. To mitigate this, we explicitly analyze fairness and generalization across different task domains and report transferability results that highlight both strengths and limits of our approach. We disclose all relevant details transparently to foster responsible use of our method.

\section*{Reproducibility statement}
We have taken several steps to ensure reproducibility of our results. Firstly, we describe the construction process of our neuron activation feature framework in \S~\ref{sec:our_method} with explicit mathematical definitions. Additionally, detailed descriptions of the model architectures, datasets, hyperparameters, and training setups are provided in \S~\ref{sec:setups} and \S~Appendix~\ref{sec:detail_of_train_eval_setup}. Further experimental details, including ablation studies and comparisons with baselines, are presented in \S~\ref{sec:analysis} and \S~Appendices \ref{sec:different_selection_strategy} to \ref{sec:IT_for_targeted_domain}. All in-domain datasets and IT datasets used in our experiments (e.g., Alpaca-GPT4, Evo-Instruct, Orca-GPT4) are publicly available, and dataset statistics are summarized in \S~Appendix~\ref{sec:detail_of_train_eval_setup}. To facilitate reproducibility, we commit to open-sourcing the implementation of NAIT, the cross-task neuron feature library, and the Alpaca-NAIT dataset upon acceptance. These resources, along with clear documentation, will allow researchers to fully reproduce and extend our experiments.

\section*{Acknowledgments}
This work was supported by National Key Research and Development Program of China (2024YFF0908200), National Natural Science Foundation of China (Grant No. 62376262), the Natural Science Foundation of Guangdong Province of China (2024A1515030166, 2025B1515020032). This work was also supported in part by the Science and Technology Development Fund of Macau SAR (Grant Nos. FDCT/0007/2024/AKP, EF2024-00185-FST), the UM and UMDF (Grant Nos. MYRG-GRG2024-00165-FST-UMDF, MYRG-GRG2025-00236-FST), the Tencent AI Lab Rhino-Bird Research Program (Grant No. EF2023-00151-FST), the Stanley Ho Medical Development Foundation (Grant No. SHMDF-AI/2026/001), and the National Natural Science Foundation of China (Grant No. 62266013). 

\bibliography{iclr2026_conference}
\bibliographystyle{iclr2026_conference}

\appendix
\appendix
\section{Limitation}
\label{sec:limitation}

\paragraph{Model Scale} Due to computational limitations, our analysis is restricted to models with fewer than 20B parameters. Future studies could evaluate this method on larger models to provide insights into its scalability and broader implications for optimizing LLMs.

\section{The Use of Large Language Models (LLMs)}
\paragraph{Polishing the Writing with a Large Language Model}
In preparing this paper, we used a large language model to refine the writing. Typical applications included: (1) enhancing grammar and style; (2) ensuring consistency of terminology; and (3) improving the quality of translations.

\section{Comparative Analysis Before and After IT}
\label{sec:it_function}
\begin{table*}[h]
\caption{Comparison of model functions before and after instruction tuning.}
\vspace{1em}
\centering
\renewcommand{\arraystretch}{1.3}
\resizebox{\textwidth}{!}{
\begin{tabular}{p{6cm}p{6cm}p{6cm}}
    \toprule
    \textbf{Function} & \textbf{Before Instruction Tuning} & \textbf{After Instruction Tuning}  \\ 
    \midrule
    Enhancing instruction-following ability & Limited to text completion & Capable of following instructions  \\ 
    Activating latent abilities & Relies on knowledge from pertaining & Activates abilities not evident in pretraining, such as logical reasoning or task decomposition \\ 
    Enhancing downstream task performance &  General knowledge capabilities & Improved downstream task capabilities \\ 
    \bottomrule
\end{tabular}
}
\label{tab:function_of_IT}
\end{table*}
The following table~\ref{tab:function_of_IT} compares the main functional differences of models before and after IT. As shown, IT significantly improves instruction-following abilities, activates latent skills such as reasoning and task decomposition, and leads to notable gains in downstream task performance. 

\paragraph{Benchmark Settings}
We evaluated our approach on various downstream tasks using their data as in-domain data (detailed dataset descriptions are in Appendix \ref{sec:detail_of_train_eval_setup}):
\begin{itemize}[label=\textbullet, topsep=1pt, itemsep=1pt, left=1pt]
    \item \textbf{\textit{Factual Knowledge}}: We evaluate the Massive Multitask Language Understanding (MMLU) dataset \citep{DBLP:conf/iclr/HendrycksBBZMSS21}, splitting it into in-domain and test sets, and report 5-shot results. To assess the generalization capability of our model, we additionally conduct experiments on MMLU-Pro \citep{DBLP:conf/nips/WangMZNCGRAHJLK24} as an OOD benchmark. 
    \item \textbf{\textit{Math Capability}}: We use the Grade School Math (GSM) dataset \citep{DBLP:journals/corr/abs-2110-14168}, divided into in-domain and test sets, and report 5-shot results under the CoT setting. To assess the generalization capability of our model, we additionally conduct experiments on SVAMP \citep{patel-etal-2021-nlp} as an OOD benchmark.
    \item \textbf{\textit{Reasoning}}: We assess the Big-Bench-Hard (BBH) dataset \citep{DBLP:conf/acl/SuzgunSSGTCCLCZ23}, with in-domain and test splits, and report 5-shot results under the CoT setting.
    \item \textbf{\textit{Multilingual Understanding}}: We evaluate the TyDiQA benchmark \citep{DBLP:journals/tacl/ClarkPNCGCK20}, a multilingual QA dataset, using in-domain and test splits, and report 5-shot results with the gold-passage setup. To assess the generalization capability of our model, we additionally conduct experiments on XQuAD \citep{Artetxe:etal:2019} as an OOD benchmark.
    \item \textbf{\textit{Coding Capability}}: We activate the coding ability using BigCodeBench \citep{DBLP:journals/corr/abs-2406-15877} and evaluate in HumanEval \citep{DBLP:journals/corr/abs-2107-03374}, reporting pass@10 results with a sampling temperature of \texttt{0.8}. To assess the generalization capability of our model, we additionally conduct experiments on MBPP  \citep{austin2021program} as an OOD benchmark. 
\end{itemize}

\section{Details of Training and Evaluation Setup}
\label{sec:detail_of_train_eval_setup}
\paragraph{Statistics of the in-domain, IT, and Test dataset} Table~\ref{tab:data_stats} summarizes the statistics of the In-domain, IT, and Test datasets.
\begin{table}[h]
\caption{\textbf{Statistics of the Number of in-domain, IT, and Test datasets Used in Experiment.}}
\vspace{1em}
\centering
\scalebox{0.6}{%
\begin{tabular}{l cc cc ccc cc c}
\toprule
\textbf{Targeted Ability$\rightarrow$} & \multicolumn{2}{c}{\makecell{Factual\\Knowledge}} & \multicolumn{2}{c}{\makecell{Mathematical\\Reasoning}} & \multicolumn{2}{c}{\makecell{Coding\\Ability}} & \multicolumn{3}{c}{\makecell{Multilingual\\Understanding}} & \multicolumn{1}{c}{\makecell{General\\Reasoning}} \\
\toprule
\textbf{DownStream Task} & \textbf{MMLU} & \textbf{MMLU-Pro} & \textbf{GSM} & \textbf{SVAMP}  & \textbf{HumanEval} & \textbf{BigCodeBench} & \textbf{MBPP} & \textbf{TydiQA} & \textbf{XQuAD} & \textbf{BBH} \\ 
\midrule
\textbf{In-domain Dataset} & 1,531 & - & 7,473 & - & - & 1,140 & - & 49,400 & - & 4,556 \\ 
\textbf{Alpaca-GPT4 IT Dataset}   &  \multicolumn{10}{c}{52,002} \\ 
\textbf{Evol-Instruct IT Dataset}   &  \multicolumn{10}{c}{70,000} \\ 
\textbf{Oral-GPT4 IT Dataset}   &  \multicolumn{10}{c}{100,000} \\ 
\textbf{Test Dataset}           & 14042 & 2800 & 1319 & 900 & 164 & - & 974 & 900 & 1,200 & 1080  \\ 
\textbf{Test few-Shot} & \multicolumn{2}{c}{5} & \multicolumn{2}{c}{8} & \multicolumn{3}{c}{0} & \multicolumn{2}{c}{1} & \multicolumn{1}{c}{1} \\ 
\bottomrule
\end{tabular}
}
\label{tab:data_stats}
\end{table}

\paragraph{Baselines Settings}
Our comparison baselines cover a variety of advanced methods for evaluating and improving IT datasets:
\begin{itemize}[label=\textbullet, topsep=1pt, itemsep=1pt, left=1pt]
    \item \textbf{\textit{Alpaca-GPT4}}~\citep{DBLP:journals/corr/abs-2304-03277}: A widely utilized IT dataset that leverages a self-instruct methodology, enabling the autonomous generation of instructions through the advanced capabilities of GPT-4.
    \item \textbf{\textit{LIMA}}~\citep{DBLP:conf/nips/ZhouLX0SMMEYYZG23}: It primarily comprises 1,000 meticulously crafted, high-quality IT data points designed to enhance the alignment capabilities of LLMs. 
    % \item \textbf{\textit{LESS}\citep{}}: 
    \item \textbf{\textit{AlpaGasus}}~\citep{DBLP:conf/iclr/ChenLYWGYTS0HJ24}: This approach employs the advanced capabilities of ChatGPT to evaluate and selectively curate data from the original Alpaca dataset.
    \item \textbf{\textit{Q2Q}}~\citep{DBLP:conf/naacl/LiZLCC0W0024}: It functions by training a precursor model and assessing the quality of the instructional data based on two distinct loss values derived from this model. 
    % \item \textbf{\textit{Instruction Mining}\citep{}}: This approach involves analyzing data features and loss values to develop a formula for evaluating data quality.
    \item \textbf{\textit{SelectIT}}~\citep{DBLP:conf/nips/Liu0W0WH024}: A novel method that leverages the intrinsic uncertainty of LLMs to select high-quality IT data without requiring external resources.
\end{itemize}

\begin{table}[t]
\centering
\caption{\textbf{Hyperparameter settings for supervised fine-tuning.}}
\vspace{0.5em}
% 定义三列：参数名，主要实验值，消融实验值
\begin{tabular}{lcc}
\toprule
\multirow{2}{*}{\textbf{Parameter Key}} & \multicolumn{2}{c}{\textbf{Value}} \\ 
\cmidrule(lr){2-3} % 画一条横线，覆盖第2到第3列
 & \textbf{Main Experiment} & \textbf{Ablation Experiment} \\
\midrule
\textbf{Learning rate} & \multicolumn{2}{c}{2.0e-5} \\ % 相同的值合并
\textbf{Cutoff length} & \multicolumn{2}{c}{2048} \\ 
\textbf{LR scheduler}  & \multicolumn{2}{c}{cosine} \\
\textbf{bf16}          & \multicolumn{2}{c}{True} \\
\textbf{Warmup ratio}  & \multicolumn{2}{c}{0.03} \\
\textbf{Weight decay}  & \multicolumn{2}{c}{0.1} \\
\cmidrule(lr){2-3} 
\textbf{Epoch}         & 3 & 1 \\
\textbf{Batch size}    & 128 & 32 (14B) / 64 (8B) \\
\bottomrule
\end{tabular}
\label{tab:sft_settings}
\end{table}

\paragraph{Fine-tuning Parameters}
The model is fine-tuned over \texttt{3} epochs with a batch size of \texttt{128} to ensure efficient learning while avoiding overfitting. The optimization process employs the Adam optimizer with hyperparameters set to $\beta_{1}=\texttt{0.9}$ and $\beta_{2}=\texttt{0.999}$, which are standard values for stable and effective training. The learning rate follows a cosine decay schedule, starting at \texttt{2e-5} and decreasing gradually to \texttt{0}, a strategy known to improve convergence by reducing oscillations during later training stages.
To maximize the model's performance, we utilize an input sequence length of \texttt{2048} tokens, as this configuration has been demonstrated to be effective for LLMs in handling long-context tasks. Table~\ref{tab:sft_settings} provides an overview of the key hyperparameters used during training on \texttt{4} \texttt{A800} GPUs with \texttt{1} node. For the main experiments, we train for 3 epochs with a batch size of 128. For ablation studies, we reduce the training to 1 epoch and adjust the batch size (32 for Llama-2-14B, 64 for Llama-3-8B) to accommodate different model sizes.

\paragraph{Text Generation Settings}
For text generation, we employ greedy search for computational efficiency, and set the temperature to \texttt{0.8} and the top-p to \texttt{0.95} following the HumanEval setting~\citep{DBLP:journals/corr/abs-2107-03374}, to enhance the creativity and diversity of the generated content while maintaining its accuracy and contextual relevance.

\section{NAIT VS. other Baseline with activating targeted ability}
\label{app:nait_vs_less}
In this section, we elaborate on the experiment setting and analysis between \textsc{Nait} and other Baseline with activating targeted ability, incuding embedding-based methods, representation method \citep{DBLP:conf/cvpr/ZhangIESW18,DBLP:conf/iclr/HanawaY0I21} the gradient-based method (LESS \citep{DBLP:conf/nips/XieS0L23}). 
\renewcommand{\arraystretch}{1.2}
\begin{table*}[t]
\caption{\textbf{Performance comparison of baselines of activating targeted ability using 10\% of the IT data.} (e.g., +MMLU refers to the process where an in-domain dataset to guide the IT data selection.) The \textbf{Bold} represents the best performance respectively in each column. }
\vspace{1em}
\centering
\scalebox{0.75}{ % 缩放比例为 0.75
\begin{tabular}{l cc cc c c}
\toprule
\textbf{Method$\downarrow$} & \multicolumn{2}{c}{\makecell{Factual\\Knowledge}} & \multicolumn{2}{c}{\makecell{Multilingual\\Understanding}} & \multicolumn{1}{c}{\makecell{General\\Reasoning}} \\
 \textbf{Test$\rightarrow$} & \textbf{MMLU} & \textbf{MMLU-Pro}  & \textbf{TydiQA} & \textbf{XQuAD}  & \textbf{BBH}  & \textbf{AVG} \\
\midrule
\multicolumn{7}{c}{\emph{+MMLU}} \\
\hdashline
\textsc{Embedding} & 46.29 & 21.07 & 42.97 & 44.73 & 36.67 & 38.35 \\
\textsc{Representation-based} & 45.57 & 23.0 & \underline{46.78} & 46.85 & 38.15 & \underline{40.07} \\
\textsc{LESS} & \textbf{48.12} & \textbf{24.50} & 43.42 & 45.51 & \underline{38.15} & 39.94 \\
\textsc{Nait} & \underline{47.81} & \underline{23.61} & \textbf{47.16} & \textbf{49.47} & \textbf{38.52} & \textbf{41.31} \\
\midrule
\multicolumn{7}{c}{\emph{+TydiQA}} \\
\hdashline
\textsc{Embedding-based} & \textbf{47.49} & \underline{22.54} & 42.4 & 46.53 & \underline{38.43} & 39.48 \\
\textsc{Representation-based} & 45.91 & 21.04 & 40.02 & 46.0 & 36.02 & 37.80 \\
\textsc{LESS} & 44.68 & 22.21 & \underline{45.24} & \underline{47.67} & 37.69 & \underline{39.50} \\
\textsc{Nait} & \underline{46.17} & \textbf{22.82} & \textbf{47.78} & \textbf{49.23} & \textbf{40.00} & \textbf{41.20} \\
\midrule
\multicolumn{7}{c}{\emph{+BBH}} \\
\hdashline
\textsc{Embedding} & \underline{47.51} & 22.86 & 43.25 & 46.66 & 36.76 & 39.41 \\
\textsc{Representation-based} & 45.81 & 21.11 & 36.78 & 44.61 & 37.59 & 37.18 \\
\textsc{LESS} & 46.85 & \textbf{23.75} & \textbf{46.56} & \underline{47.67} & \underline{39.72} & \underline{39.61} \\
\textsc{Nait} & \textbf{47.78} & \underline{23.36} & \underline{45.93} & \textbf{48.46} & \textbf{40.46} & \textbf{41.20} \\
\midrule
\end{tabular}
}
\vspace{-5mm}
\label{table:nait_vs_less}
\end{table*}
\paragraph{Experiment Setting}
\begin{itemize}[label=\textbullet, topsep=1pt, itemsep=1pt, left=1pt]
    \item \textbf{\textit{Embedding-based Method}}: We utilize the base model (Llama-2-7b-hf) to extract the static input embeddings. Specifically, we compute the mean of the embeddings across all tokens in the sequence to obtain a global vector representation for each data sample.
    \item \textbf{\textit{Representation-based Method}}~\citep{DBLP:conf/cvpr/ZhangIESW18,DBLP:conf/iclr/HanawaY0I21}:  We extract the hidden states of all tokens from the final decoder layer and apply mean pooling to derive a deep semantic representation of the input.
    \item \textbf{\textit{LESS}}~\citep{DBLP:conf/nips/XieS0L23}: As a representative gradient-based approach, LESS selects samples by estimating their influence on the target task via low-rank gradient embeddings. We utilize conversation format of Alpaca-GPT4\footnote{https://huggingface.co/datasets/liangxin/Alpaca\_GPT4} as the candidate pool. Following the standard of our experiment, we first perform a 3 epoch LoRA warmup to obtain feature checkpoints. Then, using BBH, TydiQA, and MMLU provided by LESS as target validation sets for gradient estimation, we compute influence scores and select the top-$k$ samples from Alpaca-GPT4 that are expected to minimize the validation loss on these targets.
\end{itemize}

\paragraph{Result Analysis}
While LESS achieves high scores on targeted tasks (e.g., 48.12 on MMLU vs. \textsc{Nait}'s 47.81), it suffers from overfitting to the target validation set. This results in poor generalization, as evidenced by its lower average score across all tasks (39.94) compared to \textsc{Nait} (41.31). \textsc{Nait} maintains robust performance across both targeted and non-targeted settings (e.g., TydiQA and BBH), demonstrating superior transferability. Embedding-based and representation methods rely on surface-level semantic similarity. While computationally cheap, they often fail to capture complex task-specific capabilities. Gradient-based methods (e.g., LESS) calculate influence functions involving gradients and Hessian approximations. This approach is computationally prohibitive and memory-intensive for large-scale models, despite its precision on specific targets.

% \section{Layer of Target Ability Analysis}

\section{Algorithmic Framework}
\label{app:alg_fra}
In this section, we present the detailed pseudo-code for our proposed method, as discussed in Section \ref{sec:method}. Algorithm outlines the complete procedure, including the extraction of neuron activation features and the subsequent activation feature-guided data selection.
\begin{algorithm}[h]
\caption{Neuron-aware Instruction-tuning Data Selection}
\label{alg:cpca_refined}
\begin{algorithmic}[1]
\Require In-domain data $\mathcal{P}$, Instruction data $\mathcal{D}_{\text{ins}}$, Model $\mathcal{M}$, Activation layers $\mathcal{L}$, Selection budget top-$k$
\Ensure Selected subset $\mathcal{D}_{\text{selected}}$

\State \textbf{Stage 1: Targeted Ability Activation Capture}
\State Initialize activation difference set $\Delta \mathcal{A}^{(l)} \gets \emptyset$ for each layer $l \in \mathcal{L}$
\For{each in-domain sample $P_i = (t_1, t_2, \dots, t_K) \in \mathcal{P}$}
    \For{each layer $l \in \mathcal{L}$}
        \State Extract activations $\mathcal{A}^{(l)}(t_1)$ and $\mathcal{A}^{(l)}(t_K)$ from $\mathcal{M}$
        \State $\Delta\mathcal{A}_i^{(l)} \gets \mathcal{A}^{(l)}(t_K) - \mathcal{A}^{(l)}(t_1)$ \Comment{Compute activation change from first to last token}
        \State $\Delta \mathcal{A}^{(l)} \gets \Delta \mathcal{A}^{(l)} \cup \{\Delta\mathcal{A}_i^{(l)}\}$
    \EndFor
\EndFor

\State \textbf{Stage 2: Direction Extraction via PCA}
\State Initialize direction vectors $\mathcal{V} \gets \emptyset$
\For{each layer $l = 1, \dots, L$}
    \State $\mathbf{v}_l \gets \text{PCA}(\Delta \mathcal{A}^{(l)})$ \Comment{Get First Principal Component}
    \State $\mu_{\text{diff}} \gets \frac{1}{|\mathcal{P}|} \sum_{i=1}^{|\mathcal{P}|} (\mathcal{A}^{(l)}(t_K) - \mathcal{A}^{(l)}(t_1))$
    \If{$\mu_{\text{diff}} \cdot \mathbf{v}_l < 0$} \Comment{Align Direction With Activation Feature}
        \State $\mathbf{v}_l \gets -\mathbf{v}_l$
    \EndIf
    \State $\mathcal{V} \gets \mathcal{V} \cup \{ \mathbf{v}_l \}$
\EndFor

\State \textbf{Stage 3: Activation Feature-guided Data Scoring}
\State Initialize scores $\mathcal{S} \gets \emptyset$
\For{each sample $y \in \mathcal{D}_{\text{ins}}$}
    \State Extract Activation $\mathcal{A}^{(l)}$ using $\mathcal{M}(y)$
    \State $s_y \gets \sum_{l=1}^{L} (\mathcal{A}^{(l)} \cdot \mathbf{v}_l)$ \Comment{Project Instruct-tuning Data Onto Target Direction}
    \State $\mathcal{S} \gets \mathcal{S} \cup \{ s_y \}$
\EndFor

\State \textbf{Stage 4: Data Ranking \& Selection}
\State $\mathcal{D}_{\text{selected}} \gets \text{top-}k(\mathcal{S}) $
\State \Return $\mathcal{D}_{\text{selected}}$

\label{alg:main}
\end{algorithmic}
\end{algorithm}

\section{The Details of Different Selection Strategies}
Table \ref{tab:different_selection_strategy} compares selection strategies using 10\% of the IT dataset. \textsc{Nait} achieves the best performance with a +3.00\% improvement over the baseline (Alpaca-GPT4), while other strategies like Hard (PPL) and Easy (Length) significantly reduce performance. This highlights \textsc{Nait}'s effectiveness and the importance of careful data selection.
\label{sec:different_selection_strategy}
\begin{table*}[h]
\caption{\textbf{Performance comparison of different selection strategies using 10\% of the IT data.} The \textbf{Bold} represents the best performance respectively in each column. \textit{$\Delta$ ($\uparrow$)} indicates the performance improvement relative to the ID 01 baseline.}
\vspace{1em}
\centering
\scalebox{0.9}{ % 调整大小
\begin{tabular}{cccccccc}
\toprule
\multirow{2}{*}{\textbf{Method}} & \multirow{2}{*}{\textbf{MMLU}} & \multirow{2}{*}{\textbf{BBH}} & \multirow{2}{*}{\textbf{GSM}} & \multirow{2}{*}{\textbf{TydiQA}} & \multirow{2}{*}{\textbf{CodeX}} & \multicolumn{2}{c}{\textbf{Overall}} \\
\cmidrule(lr){7-8}
& & & & & & \textbf{AVG} & \textit{$\Delta$ ($\uparrow$)} \\
\midrule
~Alpaca-GPT4     &  \bf{46.87} & 39.94 & 14.63 & 39.48  & 27.87 & 34.16 & -  \\
~+ Hard (PPL)    & 45.52 & 22.22 & 8.72  & 27.87  & 20.56 & 24.98 & -26.87\%  \\
~+ Easy (PPL)    & 46.78 & 38.15 & 15.39 & 33.12  & 28.93 & 32.47 & -5.53\% \\
~+ Hard (Length) & 46.10 & 36.94 & 16.00 & 34.32  & \bf{30.89} & 32.85 & -3.83\% \\
~+ Easy (Length) & 44.15 & 10.28 & 6.37  & 28.07  & 17.00 & 21.17 & -38.03\% \\
~+ \textsc{Nait}          &  46.83 & \bf{40.02} & \bf{16.53} & \bf{46.09}  & 26.44 & \bf{35.18} & \bf{+3.00\%} \\
\bottomrule
\end{tabular}
}
\vspace{-1em}
\label{tab:different_selection_strategy}
\end{table*}

\section{NAIT for cross-lingual}
\label{sec:IT_for_targeted_domain}
\begin{table*}[h]
\caption{\textbf{Overall results on machine translation with LLMs using the ALMA-7b.} \textsc{Nait} indicates selecting the top 30\% of data by \textsc{Nait}, Rand. refers to randomly selecting 30\% of data, and Full uses the full dataset.
}
\vspace{1em}
\scalebox{0.8}{
\begin{tabular}{l cc cc cc cc cc}
\toprule
\multirow{2}{*}{\bf Method} 
    & \multicolumn{2}{c}{\bf En$\Rightarrow$De} 
    & \multicolumn{2}{c}{\bf De$\Rightarrow$En} 
    & \multicolumn{2}{c}{\bf Zh$\Rightarrow$En} 
    & \multicolumn{2}{c}{\bf En$\Rightarrow$Zh} 
    & \multicolumn{2}{c}{\bf Overall} \\
\cmidrule(lr){2-3}
\cmidrule(lr){4-5}
\cmidrule(lr){6-7}
\cmidrule(lr){8-9}
\cmidrule(lr){10-11}
    & COMET & BLEU 
    & COMET & BLEU 
    & COMET & BLEU 
    & COMET & BLEU 
    & COMET & BLEU \\
\midrule
% \multicolumn{11}{c}{\it Our Implemented Method} \\
+ Full          & 84.54 & 28.24 & \bf{83.13} & \bf{29.06} & 79.80 & 22.89 & \bf{84.15} & 34.58 & 82.91 & 28.69 \\
+ Rand.         & 82.44 & 28.29 & 81.12 & 27.20 & 78.53 & 21.11 & 82.12 & 33.91 & 81.05 & 27.63 \\
+ \textsc{Nait}          & \bf{85.14} & \bf{29.02} & 82.50 & 28.70 & \bf{80.22} & \bf{23.11} & 83.89 & \bf{35.92} & \textbf{82.94} & \textbf{29.19} \\
\bottomrule
\end{tabular}
}
\label{fig:MT_LLMs_apd}
\end{table*}

As shown in Table \ref{fig:MT_LLMs_apd}, we further evaluate the cross-lingual data selection capability of \textsc{Nait} on the multilingual IT dataset in the context of machine translation. We adopt the powerful ALMA-7b \citep{DBLP:conf/iclr/Xu0SA24} model as the backbone and conduct experiments on four representative translation directions: English-German, German-English, Chinese-English, and English-Chinese. The experimental results demonstrate that the \textsc{Nait} method outperforms both random selection (Rand.) and using the full dataset (Full) on most evaluation metrics. Notably, \textsc{Nait} achieves the highest overall COMET (82.94) and BLEU (29.19) scores, further highlighting its superior generalization ability and robustness in both target domains and cross-lingual scenarios.

\section{case study}
To intuitively demonstrate the effectiveness of our method, we provide a qualitative comparison of the training samples selected by Embedding, LESS, and \textsc{Nait} for different downstream benchmarks in Table~\ref{tab:compare_example}
As observed in the table, baseline methods often select data based on surface-level textual similarity, which may not align with the complexity of the target task. For instance, in TydiQA, the embedding method selects translation tasks but may not align with the required question answering. For the general reasoning ability, the Embedding and LESS strategies select simple tasks such as word frequency counting or text summarization. In contrast, \textsc{Nait} identifies an algorithmic problem that requires multi-step logical reasoning and coding synthesis. Similarly, for MMLU, \textsc{Nait} selects a knowledge-style question similar to other methods that aligns well with the knowledge-intensive of the benchmark. 
\label{app:case_study}\begin{table*}[t]
% \vspace{-12em}
\caption{Qualitative comparison of examples selected by Embedding, LESS, and NAIT strategies across different benchmarks. The samples here from \textsc{Nait} are outside of the intersection of general core data.}
\vspace{1em}
\centering
\scriptsize
\begin{tabular}{p{0.3\textwidth} p{0.3\textwidth} p{0.3\textwidth}}
\toprule
\multicolumn{3}{c}{\textbf{Targeted ability activated by TydiQA}} \\
\midrule
\multicolumn{1}{c}{\textbf{Selected by Embedding}} &
\multicolumn{1}{c}{\textbf{Selected by LESS}} & 
\multicolumn{1}{c}{\textbf{Selected by NAIT}} \\ 
\midrule
\textbf{Instruction:} Translate the phrase "Yechh" into Hindi.\newline \newline \textbf{Output:} The phrase ``Yechh'' is an expression of disgust or dislike and is not a word with a direct translation into Hindi. However, a similar expression of disgust in Hindi would be ``{\dn .phahaH}'' or ``{\dn chii}'' (Chhee) or ``{\dn caDham}'' (chidam) or ``{\dn are baapa re}'' (arey baap re).  &
\textbf{Instruction:} Construct a mathematical equation for the following problem:\newline  At a party, there are 3 types of food. Each type of food is served three times each. Guests can choose only 1 of each food serving. How many different combinations are there for the guests?\newline \textbf{Output:} number of combinations = 3 x 3 x 3 = 27 &
\textbf{Instruction:} Create two authentic Thai dishes. \newline  \textbf{Output:} Great! Here are the two Thai dishes!\newline 1. Thai Green Curry Chicken\newline Ingredients:\newline - 1 1/2 lbs boneless, skinless chicken breasts\newline - 1 can (13/14-ounce) coconut milk\newline - 1/4 cup green curry paste\newline [...] \\
\bottomrule
\multicolumn{3}{c}{\textbf{Targeted ability activated by  BBH}} \\
\midrule
\multicolumn{1}{c}{\textbf{Selected by Embedding}} &
\multicolumn{1}{c}{\textbf{Selected by LESS}} & 
\multicolumn{1}{c}{\textbf{Selected by NAIT}} \\ 
\midrule
\textbf{Instruction:} Given a document, find the most frequently used words. \newline  \textbf{Input:} John went to the store to buy some candy.\newline \textbf{Output:} Based on the given sentence, the most frequently used words are:\newline - "to" - which appears 2 times in the sentence. \newline - "the" - which appears 1 time in the sentence.\newline - "John" - which appears 1 time in the sentence.\newline - "went" - which appears 1 time in the sentence.\newline - "store" - which appears 1 time in the sentence.\newline - "buy" - which appears 1 time in the sentence.\newline - "some" - which appears 1 time in the sentence.\newline - "candy" - which appears 1 time in the sentence. &
\textbf{Instruction:} Convert the following document into a list of five bullet points. \newline  \textbf{Input:} Sports play a key role in the lives of many people. It can teach young people about leadership and teamwork, as well as important values like perseverance and camaraderie. Sporting events bring communities together and also provide an opportunity for people to come and support their favourite teams or players.\newline  \textbf{Output:} - Sports play a key role in the lives of many people\newline - Sports can teach young people about leadership and teamwork, as well as important values like perseverance and camaraderie.\newline - Sporting events bring communities together. \newline - Sporting events provide an opportunity for people to come and support their favourite teams or players.\newline - Sports promote important values and bring people together. &
\textbf{Instruction:} Design an algorithm to solve the Eight Queens puzzle \newline \textbf{Output:} The Eight Queens puzzle is a classic problem where the goal is to place eight queens on an 8x8 chessboard in such a way that no two queens threaten each other. This means that no two queens can be placed in the same row, column, or diagonal.\newline One solution to this problem is to use backtracking. Here is an algorithm to solve the Eight Queens puzzle using backtracking:\newline 1. Start with an empty 8x8 chessboard and add the first queen to the first row in the first column.\newline 2. Move to the next row and try placing a queen in each column. Check if the position is safe by verifying if any of the previously placed queens is attacking the current position. If it’s a safe position, place a queen in that column.\newline 3.Boolean function is safe(board, row, col):\newline [...] \\
\bottomrule
\multicolumn{3}{c}{\textbf{Targeted ability activated by MMLU}} \\
\midrule
\multicolumn{1}{c}{\textbf{Selected by Embedding}} &
\multicolumn{1}{c}{\textbf{Selected by LESS}} & 
\multicolumn{1}{c}{\textbf{Selected by NAIT}} \\ 
\midrule
\textbf{Instruction:} Rank the following four choices in order from least likely to most likely.\newline \textbf{Input:} A. Taking a long walk\newline B. Eating two slices of chocolate cake\newline C. Writing a persuasive essay\newline D. Drinking a glass of orange juice\newline \textbf{Output:} It's not possible to rank the given choices in order from least likely to most likely without additional context or information about the person's habits, preferences or circumstances. Could you provide more information so I can better assist you? &
\textbf{Instruction:} Choose the best answer from the given list of choices.\newline  \textbf{Input:} Question: Which of the following is a benefit of cloud computing?\newline A. Lower cost\newline B. Reliability\newline C. Increased scalability\newline D. All of the above\newline \textbf{Output:} D. All of the above. & 
\textbf{Instruction:} Generate an example of a multiple choice question related to the following sentence.\newline  
\textbf{Output:} Entomology is the science which deals with:\newline 
A) human behavior\newline 
B) the study of insects\newline 
C) marine life\newline 
D) the solar system\newline 
Correct answer: B \\
\bottomrule
\end{tabular}
\label{tab:compare_example}
\end{table*}

\end{document}